\documentclass[10pt,twocolumn]{article}

\usepackage{amsmath,amsfonts,bm}

\def\eqref#1{(\ref{#1})}
\def\1{\bm{1}}

\DeclareMathAlphabet{\mathsfit}{\encodingdefault}{\sfdefault}{m}{sl}
\SetMathAlphabet{\mathsfit}{bold}{\encodingdefault}{\sfdefault}{bx}{n}

\usepackage{cvpr}
\usepackage{times}
\usepackage{epsfig}
\usepackage{graphicx}
\usepackage{amsmath}
\usepackage{amssymb}
\usepackage{multirow}
\usepackage{footmisc}
\usepackage{color}
\usepackage[dvipsnames]{xcolor}
\usepackage{colortbl,booktabs}
\usepackage{threeparttable}
\usepackage{nopageno}
\usepackage{comment}
\usepackage[small,bf]{caption}
\usepackage{url}
\usepackage{algorithm}
\usepackage{algpseudocode}
\usepackage{mathtools}
\usepackage{pifont}
\usepackage{lipsum}
\usepackage{color,xcolor}
\definecolor{Sijia_color}{rgb}{0.858, 0.188, 0.478}
\definecolor{Tianlong_color}{rgb}{0.00, 0.00, 1.00}
\usepackage{adjustbox}
\usepackage{booktabs}       % professional-quality tables
\usepackage{amsfonts}       % blackboard math symbols
\usepackage{nicefrac}       % compact symbols for 1/2, etc.
\usepackage{microtype}      % microtypography
\usepackage{makecell}
\usepackage{diagbox}

\DeclarePairedDelimiterX{\inp}[2]{\langle}{\rangle}{#1, #2}

\DeclareMathOperator*{\minimize}{\text{minimize}}
\DeclareMathOperator*{\maximize}{\text{maximize}}

\DeclareMathAlphabet\mathbfcal{OMS}{cmsy}{b}{n}
\newcommand{\Def}[0]{\mathrel{\mathop:}=}

\newcommand{\SL}[1]{\textcolor{Sijia_color}{#1}}
\newcommand{\TL}[1]{\textcolor{Tianlong_color}{TL: #1}}

\usepackage[pagebackref=true,breaklinks=true,colorlinks,bookmarks=false]{hyperref}

\cvprfinalcopy % *** Uncomment this line for the final submission

\def\cvprPaperID{5580} % *** Enter the CVPR Paper ID here

\ifcvprfinal\fi
\begin{document}

%%%%%%%%% TITLE
%\title{Self-Supervised Pretraining Meets Adversarial Robustness: \\ Towards Pretraining that Leads to Robust Models}

\title{Adversarial Robustness: From Self-Supervised Pre-Training  to Fine-Tuning}

\author{Tianlong Chen\textsuperscript{1}, Sijia Liu\textsuperscript{2}, Shiyu Chang\textsuperscript{2}, Yu Cheng\textsuperscript{3}, Lisa Amini\textsuperscript{2}, Zhangyang Wang\textsuperscript{1}\\
\textsuperscript{1}Texas A\&M University, \textsuperscript{2}MIT-IBM Watson AI Lab, IBM Research \textsuperscript{3}Microsoft Dynamics 365 AI Research\\
\textit{\small \{wiwjp619,atlaswang\}@tamu.edu, \{sijia.liu,shiyu.chang,lisa.amini\}@ibm.com, yu.cheng@microsoft.com} %\\
% \url{https://github.com/TAMU-VITA/Adv-SS-Pretraining}
}

% \author{Tianlong Chen\\
% Texas A\&M University\\
% {\tt\small wiwjp619@tamu.edu}
% \and
% Sijia Liu\\
% MIT-IBM Watson AI Lab, IBM Research	\\
% {\tt\small sijia.liu@ibm.com}
% \and
% Shiyu Chang\\
% MIT-IBM Watson AI Lab, IBM Research	\\
% {\tt\small shiyu.chang@ibm.com}
% \and
% Yu Cheng\\
% Microsoft Dynamics 365 AI Research	\\
% {\tt\small yu.cheng@microsoft.com}
% \and
% Lisa Amini\\
% MIT-IBM Watson AI Lab, IBM Research	\\
% {\tt\small lisa.amini@us.ibm.com}
% \and
% Zhangyang Wang\\
% Texas A\&M University\\
% {\tt\small atlaswang@tamu.edu}
% }

\maketitle
%\thispagestyle{empty}

%%%%%%%%% ABSTRACT
\begin{abstract}
Pretrained models from self-supervision are prevalently used in fine-tuning downstream tasks faster or for better accuracy. However, gaining robustness from pretraining is left unexplored. We introduce adversarial training into self-supervision, to provide general-purpose robust pretrained models for the first time. We find these robust pretrained models can benefit the subsequent fine-tuning in two ways: \textbf{i) boosting final model robustness; ii) saving the computation cost, if proceeding towards adversarial fine-tuning}. We conduct extensive experiments to demonstrate that the proposed framework achieves large performance margins (\eg, $3.83$\% on robust accuracy and $1.3$\% on standard accuracy, on the CIFAR-10 dataset), compared with the conventional end-to-end adversarial training baseline. Moreover, we find that different self-supervised pretrained models have diverse adversarial vulnerability. It inspires us to ensemble several pretraining tasks, which boosts robustness more. Our ensemble strategy contributes to a further improvement of $3.59$\% on robust accuracy, while maintaining a slightly higher standard accuracy on CIFAR-10. Our codes are available at \url{https://github.com/TAMU-VITA/Adv-SS-Pretraining}.
% \Lisa{Lisa: You can use this color.}

% Embeddings from self-supervision are prevalently used in fine-tuning downstream tasks. However, gaining robustness from pretrained embeddings left unexplored. We introduce adversarial training into self-supervision, to provide general-purpose robust pretrained embeddings for the first time. We find these robust pretrained embeddings can benefit fine-tuned models in two ways: \textbf{i) boosting final model robustness; ii) saving the computation cost, if proceeding towards adversarial fine-tuning}. We conduct extensive experiments on the CIFAR-10 dataset to demonstrate that the proposed framework achieves a large performance margin ($3.83$\% on robust accuracy and $1.3$\% on standard accuracy), compared with the \SL{end-to-end} adversarial training baseline. Moreover, we find that different self-supervised pretrained embeddings have a diverse adversarial vulnerability. It inspires us to ensemble pretrained embeddings, which boosts robustness more. Our ensemble strategy further contributes to an improvement of $3.59$\% on robust accuracy, while maintaining a slightly higher standard accuracy.
\end{abstract}

%%%%%%%%% BODY TEXT
\vspace{-5mm}

% \section*{Dedication}
% \vspace{-0.5em}
% \noindent Zhangyang and Shiyu would like to dedicate the paper to \textbf{Mrs Margaret Huang} (05/02/1934 - 01/26/2020), for all her love and care.

\section{Introduction}
\vspace{-0.3em}
Supervised training of deep neural networks requires massive, labeled datasets, which may be unavailable and costly to assemble \cite{hinton2006fast,bengio2007greedy,raina2007self,vincent2010stacked}. Self-supervised and unsupervised training techniques attempt to address this challenge by eliminating the need for manually labeled data. Representations pretrained through self-supervised techniques enable fast fine-tuning to multiple downstream tasks, and lead to better generalization and calibration  \cite{liu2019towards,mohseni2020self}. Examples of tasks proven to attain high accuracy through self-supervised pretraining include position predicting tasks (\textit{Selfie} \cite{trinh2019selfie}, \textit{Jigsaw} \cite{noroozi2016unsupervised, carlucci2019domain}), rotation predicting tasks (\textit{Rotation} \cite{gidaris2018unsupervised}), and a variety of other perception tasks \cite{criminisi2004region,zhang2016colorful,dosovitskiy2015discriminative}.

\begin{figure}[t]
\begin{center}
   \includegraphics[width=1\linewidth]{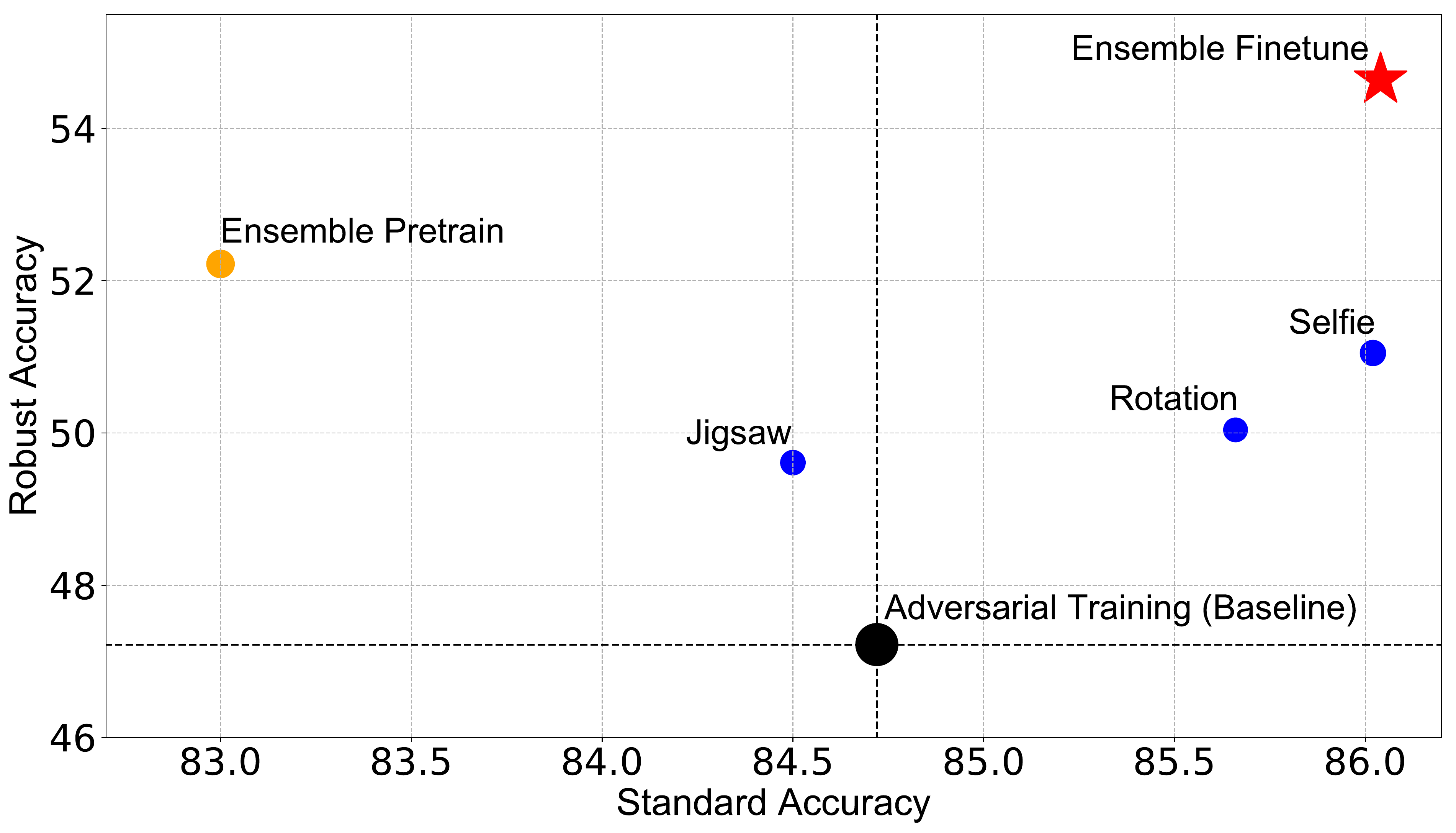}
\end{center}
\vspace{-5mm}
   \caption{Summary of our achieved performance  (CIFAR-10). \textbf{The upper right corner indicates the best performance in terms of both standard and robust accuracy.} The \textit{size of markers} represents the number of training epochs to achieve the best robust accuracy. \textit{Black circle} (\textcolor{black}{$\bullet$}) is the baseline method: \textit{end-to-end adversarial training}. \textit{Blue circles} (\textcolor{blue}{$\bullet$}) are fine-tuned models that inherit robust models from different self-supervised pretraining tasks. \textit{Orange circle} (\textcolor{orange}{$\bullet$}) is the ensemble of three self-supervised pretraining tasks. \textit{Red Star} (\textcolor{red}{\ding{72}}) is the ensemble of three fine-tuned models. {The correspondence between the  $\mathrm{marker\ size}$  and \# epochs is given by, \textit{Ensemble Fine-tune} (\textcolor{red}{\ding{72}}, 144 epochs) $>$ \textit{Baseline} (\textcolor{black}{$\bullet$}, 99 epochs) $>$ \textit{Ensemble Pretrain} (\textcolor{orange}{$\bullet$}, 56 epochs) $>$ \textit{Selfie} (\textcolor{blue}{$\bullet$}, 50 epochs) $>$ \textit{Jigsaw} (\textcolor{blue}{$\bullet$}, 48 epochs) $>$ \textit{Rotation} (\textcolor{blue}{$\bullet$}, 46 epochs)}
   }
\label{fig:intro}
%\vspace{-5mm}
\end{figure}

The labeling and sample efficiency challenges of deep learning are further exacerbated by vulnerability to adversarial attacks. For example, Convolutional Neural Networks (CNNs), are widely leveraged for perception tasks, due to high predictive accuracy.  However, even a well-trained CNN suffers from high misclassification rates when imperceivable perturbations are applied the input~\cite{Kurakin2016AdversarialML, moosavi2016deepfool}. As suggested by \cite{schmidt2018adversarially},  the sample complexity of learning an adversarially robust model with current methods is significantly higher than that of standard learning. Adversarial training (AT) \cite{madry2017towards}, the state-of-the-art model defense approach, is also known to be computationally more expensive than standard training (ST). The above facts make it especially meaningful to explore:
\begin{center}
	\begin{minipage}{0.42\textwidth}
		\it 
	\indent Can appropriately pretrained models play \textbf{a similar role} for adversarial training as they have for ST? That is, can they lead to more efficient fine-tuning and better, adversarially-robust generalization?
	\end{minipage}
\end{center}

%\SL{The transition to adv. robustness is not smooth. Better motivate robustness first.}

\iffalse
Recently, a magnitude of popular self-supervised learning tasks are proposed to provide good pretrained models, such as position predicting tasks (\textit{Selfie} \cite{trinh2019selfie}, \textit{Jigsaw} \cite{noroozi2016unsupervised, carlucci2019domain}), rotation predicting tasks (\textit{Rotation} \cite{gidaris2018unsupervised}), etc \cite{criminisi2004region,zhang2016colorful,dosovitskiy2015discriminative}. %\SL{why is the above paragraph here?}
\fi 
Self-supervision has only recently been linked to the study of robustness.
%It was not until very recently that self supervision is linked to the study of robustness. 
%As self-supervised tasks are demonstrated to help the prediction accuracies of deep supervised classifiers, it remains yet to be explored how they might affect their adversarial robustness. 
An approach is offered in \cite{hendrycks2019using}, by %treating 
incorporating
the self-supervised task as a complementary objective, which is co-optimized with the conventional classification loss through the method of AT \cite{madry2017towards}. Their co-optimization approach presents scalability challenges, and does not enjoy the benefits of pretrained embeddings.  Further, it leaves many unanswered questions, especially with respect to efficient tuning, which we tackle in this paper.
%and jointly optimizing the robustness and the auxiliary losses. 
%ZW: 不要说人家方法不行，说我们能做到universal embedding for downstream tasks!

%offers a possible solution. They use jointly optimizing framework which treats self-supervised learning task as a separate auxiliary head. Unfortunately, this approach still suffer the scalability issue and can't enjoy the benefits of universal pretrain embeddings.
%It remains challenging of how to use self-supervised learning tasks to boost adversarial robustness of deep neural networks. 

%\SL{[The motivation on why one should care about the robustness issue on pretraining + fine-tuning is not sufficient, thus, transition to the following questions and answers are too sharp to understand.]}

\paragraph{Contributions.} This paper introduces a framework for self-supervised pretraining and fine-tuning into the adversarial robustness field. We motivate our study with the following three scientific questions: 
%\vspace{-1em}
\begin{itemize}
\item [\textit{Q1:}] \textit{Is an adversarially pretrained model effective in boosting the robustness of subsequent fine-tuning?}
\vspace{-0.1em}
\item [\textit{Q2:}]\textit{Which provides the better accuracy and efficiency: adversarial pretraining or adversarial fine-tuning?}
\vspace{-0.1em}
\item [\textit{Q3:}]\textit{How does the type of self-supervised pretraining task affect the final model's robustness?}
\end{itemize}

Our contributions address the above questions and can be summarized as follows:
%\SL{[Better to make questions and answers separate. Questions belong to the part of why one should care about this research. Answers belong to your contributions.]}
\begin{itemize}
\item [\textit{A1:}] We demonstrate for the first time that 
%performing adversarial fine-tuning based on robust pretraining embeddings 
robust pretrained models leveraged for adversarial fine-tuning
result in a \textbf{large performance gain}. As illustrated by Figure~\ref{fig:intro}, the best pretrained model from a single self-supervised task (\textit{Selfie}) leads to \textbf{3.83\%} on robust accuracy\footnote{Throughout this paper, we follow \cite{zhang2019theoretically} to adopt their defined standard accuracy and robust accuracy, as two metrics to evaluate our method's effectiveness: a desired model shall be high in both.} and \textbf{1.3\%} on standard accuracy on CIFAR-10 when being adversarially fine-tuned, compared with the strong AT baseline. Even performing standard fine-tuning (which consumes fewer resources) with the robust pretrained models improves the resulting model's robustness.% to some extent. 
%$3.83$ \SL{[Confusing. $3.83\%$ improvement? compared to what baseline? Why you care about this baseline? Baseline itself (commonly-used method prior to your work) should also be mentioned before the contribution part.]} points on robust accuracy and $1.3$ points on standard accuracy. As shown in Figure~\ref{fig:intro}, all blue circles \SL{Do not use `blue circles' directly when you present contributions. Say their settings directly, e.g., fine-tuning  over a robust embedding  (learnt from a pretraining task, shown by a `blue circle' in Figure~\ref{fig:intro}) yields a substantial improvement in robust accuracy compared to AT. This comment applies to all other answers.]} have a noteworthy gap to baseline in terms of robust accuracy. Even standard fine-tuning classifier with fixed pretraining embeddings can also work to some extent.}
\vspace{-0.1em}
\item [\textit{A2:}]We systematically study all possible combinations between pretraining and fine-tuning. Our extensive results reveal that adversarial fine-tuning contributes to the dominant portion of \textbf{robustness improvement}, while robust pretraining mainly \textbf{speeds up} adversarial fine-tuning. That can also be read from Figure~\ref{fig:intro} (smaller marker sizes denote less training epochs needed). %As shown in Figure~\ref{fig:intro}, all blue and red circle have a smaller size than the baseline (black circle).
\vspace{-0.1em}
\item [\textit{A3:}] We experimentally show that the pretrained models resulting from different self-supervised tasks have diverse adversarial vulnerabilities. In view of that, we propose to pretrain with an ensemble of self-supervised tasks, in order to leverage their complementary strengths. On CIFAR-10, our ensemble strategy further contributes to an improvement of 3.59\% on robust accuracy, while maintaining a slightly higher standard accuracy. Our approach establishes a \textbf{new benchmark result} on standard accuracy ($86.04\%$) and  robust accuracy ($54.64\%$) in the setting of AT. 
%With sufficient experiments, we find there is a weak transferability of adversarial examples among fine-tuned models with different self-supervised pretraining feature embeddings. It suggests a significant robustness diversity among different self-supervised tasks. As shown in Figure~\ref{fig:intro}, \textit{Selfie} pretraining task provides a much better robust embedding, compare with other two blue circles. Due to a high diversity or inconsistency of adversarial examples, we smartly \SL{poor writing} ensemble all fine-tuned models (red star in Figure~\ref{fig:intro}) and achieve state-of-the-art results on CIFAR-10 dataset (86.04 on standard accuracy and 54.64 on robust accuracy). \SL{How about saying   `a  new benchmark on standard accuracy ($86.04\%$) and  robust accuracy ($54.64\%$) in the   setting of AT'}
\end{itemize}

\section{Related Work}
\vspace{-0.1em}
\paragraph{Self-supervised pretraining.} Numerous self-supervised learning methods have been developed in recent years, including: region/component filling (\eg \textit{inpainting} \cite{criminisi2004region} and \textit{colorization} \cite{zhang2016colorful}); rotation prediction  \cite{gidaris2018unsupervised}; category prediction \cite{dosovitskiy2015discriminative}; and patch-base spatial composition prediction (\eg, \textit{Jigsaw} \cite{noroozi2016unsupervised, carlucci2019domain} and \textit{Selfie} \cite{trinh2019selfie}). All perform standard training, and do not tackle adversarial robustness. For example, \textit{Selfie} \cite{trinh2019selfie}, generalizes BERT to image domains. It masks out a few patches in an image, and then attempts to classify a right patch to reconstruct the original image. Selfie is first pretrained on unlabeled data and fine-tuned towards the downstream classification task. 

%iv) predicting relative positions of patches (e.g. \textit{Jigsaw} \cite{noroozi2016unsupervised, carlucci2019domain} and \textit{Selfie} \cite{trinh2019selfie}). 

%In this paper, we only consider self-supervised learning tasks \SL{[why?  not clear].} (\textit{Selfie}, \textit{Rotation}, \textit{Jigsaw}) belong to classification tasks and have dominant position of performance.

%The recent work \cite{trinh2019selfie} proposed \textit{Selfie}, self-supervised pretraining for image embedding. Selfie generalizes BERT to image domains. Spurred by BERT, Selfie masks out a few patches in an image and attempts to reconstruct the original image. To enable the classification loss, Selfie asks the model to classify the right patch to fill in a target masked position. Selfie contains two stages: pretrain the model on unlabeled data and fine-tune on the target supervised task. Spurred by Selfie, we aim to learn a \textit{robust} self-supervised pretraining model that a) is independent on labels, b) preserves robust feature representations to adversarial perturbations. Since the pretrained model can be learned offline and provides a good initialization, the task-specific fine-tuning (partial or full) could significantly save the computation cost of adversarial training. 
\vspace{-1em}
\paragraph{Adversarial robustness.} Many defense methods have been proposed to improve model robustness against adversarial attacks. Approaches range from adding stochasticity~\cite{dhillon2018stochastic}, to label smoothening and feature squeezing~\cite{papernot2017extending,xu2017feature}, to denoising and training on adversarial examples ~\cite{meng2017magnet,liao2018defense}. A handful of recent works point out that those empirical defenses could still be easily compromised~\cite{athalye2018obfuscated}. Adversarial training (AT) \cite{madry2017towards} provides one of the strongest current defenses, by training the model over the adversarially perturbed training data, and has not yet been fully compromised by new attacks. \cite{gui2019model,Hu2020Triple} showed AT is also effective in compressing or accelerating models \cite{Zhu2020FreeLB} while preserving learned robustness. 

%A handful of recent works pointed out that those empirical defenses could still be easily compromised~\cite{athalye2018obfuscated}, and a few certified defenses were introduced~\cite{madry2017towards, sinha2017certifiable}. 

Several works have demonstrated model ensembles \cite{strauss2017ensemble,tramer2017ensemble} to boost adversarial robustness, as the ensemble diversity can challenge the transferability of adversarial examples. 
%One significant reason for robustness gain from the defended ensemble is the beneficial diversity among models, which potentially can generate more transferable and stronger adversarial perturbation.
Recent proposals \cite{pang2019improving,wang2019unified} formulate the diversity as a training regularizer for improved ensemble defense. Their success inspires our ensembled self-supervised pretraining.

%In this paper, we propose our diversity regularizer and apply it to adversarial training by leveraging the ensemble of multiple self-supervised learning tasks. 

%A magnitude approaches \cite{KurakinGB2016adversarial,xu2017feature,song2017pixeldefend,liao2018defense} have been proposed to defense adversarial images which are maliciously constructed by attackers. \cite{madry2017towards} propose adversarial training which is a strong defense approach that has so far not been fully compromised. It argument the training datasets with adversarial noise. Several other recent works focus on exploiting model ensemble \cite{strauss2017ensemble,tramer2017ensemble} to boost adversarial robustness. One significant reason for robustness gain from the defended ensemble is the beneficial diversity among models, which potentially can generate more transferable and stronger adversarial perturbation. \cite{pang2019improving,wang2019unified} formula the diversity into a regularizer to further improve the powerfulness of ensemble defense. In this paper, we propose our diversity regularizer and apply it to adversarial training by leveraging the ensemble of multiple self-supervised learning tasks.

\vspace{-1em}
\paragraph{Unlabeled data for adversarial robustness.} 
Self-supervised training learns effective representations for improving performance on downstream tasks, without requiring labels. Because robust training methods have higher sample complexity, there has been significant recent attention on how to effectively utilize unlabeled data to train robust models. 
%Meanwhile, due to the higher sample complexity demanded by robustness, how to utilize widely available unlabeled data to boost robustness has drawn recent interest.
%Unlabeled data hold great promise for improving representations when labeled data are scarce \cite{trinh2019selfie}. Moreover, it was shown in \cite{schmidt2018adversarially} that training models to be robust to adversarial attacks require substantially more data than those required for the standard classification task. \TL{NIPS two papers}

Results show that unlabeled data can become a competitive alternative to labeled data for training adversarially robust models. These results are concurred by \cite{zhai2019adversarially}, who also finds that learning with more unlabeled data can result in better adversarially robust generalization. 
%In our experiments, robust feature representation appears weak transferability from pretraining to fine-tuning, which suggests adversarial fine-tuning is essential. 
Both works \cite{stanforth2019labels,carmon2019unlabeled} use unlabeled data to form an \textit{unsupervised} auxiliary loss (\eg, a label-independent robust regularizer or a pseudo-label loss). 
%that is given by either a robust regularizer that has no dependence on labels or an adversarial loss that relies on pseudo-label obtained from prediction. 

To the best of our knowledge, \cite{hendrycks2019using} is the only work so far that utilizes unlabeled data via \textit{self-supervision} to train a robust model given a target supervised classification task. It improves AT by leveraging the rotation prediction self-supervision as an auxiliary task, which is co-optimized with the conventional AT loss. Our self-supervised pretraining and fine-tuning differ from all above settings.

%Although the existing work \cite{stanforth2019labels,carmon2019unlabeled,hendrycks2019using} demonstrated the potential of learning with unlabeled data to improve network robustness, they were restricted to the jointly optimizing framework over a supervised classification loss as well as an unsupervised loss that has no dependence on labels. Such an alternative to the standard adversarial training also suffers the scalability issue and can't enjoy the benefits of universal image embeddings. In our pretraining and fine-tuning design, the self-supervised pretraining offers a robust weight initialization to the task-specific fine-tuning, which not only increases the robustness gain but also speeds up the adversarial training in the fine-tuning phase.

\section{Our Proposal}
%\vspace{-0.1em}
In this section, we introduce  \textit{self-supervised pretraining} to learn feature representations from unlabeled data, followed by   \textit{fine-tuning} on a target supervised task. We  then   generalize adversarial training  (AT)  to different self-supervised pretraining and fine-tuning schemes.
%We aim to examine   how  the network robustness   is evolved  from a self-supervised pretraining stage (built on unlabeled data) to a {fine-tuning} classification  stage (built on labeled data). 

 \subsection{Setup}\label{sec: set_up}

\paragraph{Self-Supervised Pretraining}
Let $\mathcal T_{\mathrm{p}}$ denote a 
{\underline{p}retraining task} %, e.g., \textit{Selfie}, \textit{Rotation} and \textit{Jigsaw} that we  will   focus  on,
and  $\mathcal D_{\mathrm{p}}$  denote the corresponding {(unlabeled)} pretraining dataset. 
The goal of self-supervised pretraining is to learn a model 
from $\mathcal D_{\mathrm{p}}$ itself  without explicit manual supervision. This is often cast as an optimization problem, in which a proposed {pretraining loss} $\ell_{\mathrm{p}}(\boldsymbol{\theta}_{\mathrm{p}}, \boldsymbol{\theta}_{\mathrm{pc}}; \mathcal D_{\mathrm{p}})$ is minimized to determine a model parameterized by $\boldsymbol{\theta}_{\mathrm{p}}$. Here $\boldsymbol{\theta}_{\mathrm{pc}}$ signifies additional   parameters \textit{customized} for a given $\mathcal T_{\mathrm{p}}$.  
In the rest of the paper, we  focus on 
the following self-supervised pretraining tasks (details on each pretraining task are provided in the supplement):

% \textit{Selfie}  \cite{trinh2019selfie},  \textit{Rotation} \cite{gidaris2018unsupervised} and \textit{Jigsaw}  \cite{noroozi2016unsupervised, carlucci2019domain}.
% \SL{Details on the formulation of each pretraining task are provided in Appendix\,xxx.}

% \textit{Selfie} \cite{trinh2019selfie}, \textit{Jigsaw} \cite{noroozi2016unsupervised, carlucci2019domain}), rotation predicting tasks (\textit{Rotation} \cite{gidaris2018unsupervised})
% \SL{[The way that you explain every pretraining loss is not clear.]}
% \SL{[I am learning toward removing the math equations, and put details in appendix.]}

\textit{Selfie} \cite{trinh2019selfie}: 
By masking out select patches in an image,  \textit{Selfie}  constructs a classification problem to determine the correct patch to be filled in the masked location.

\textit{Rotation} \cite{gidaris2018unsupervised}: %\SL{Please write in the way that I wrote for Selfie.}
By rotating an image by a random multiple of 90 degrees, \textit{Rotation} constructs a classification problem to determine the degree of rotation applied to an input image.

\begin{comment}
\begin{align}\label{eq: rotation}
    \begin{array}{l}
\displaystyle\minimize_{\boldsymbol{\theta}} ~\mathbb E_{\mathbf x \in \mathcal D_\mathrm{p}} \left [\ell_{\mathrm{CE}}(R_{r}(\boldsymbol{x}), r\in\mathcal{G};\boldsymbol{\theta}) \right ], 
    \end{array}
\end{align}
where $R_r(\boldsymbol{x})$ is a rotation transformation; $\ell_{\mathrm{CE}}(R_{r}(\boldsymbol{x}), r\in\mathcal{G};\boldsymbol{\theta})$ is the cross-entropy between the rotation output and the ground-truth label $r$ random chosen from $\mathcal{G}=\{0^{\circ},90^{\circ},180^{\circ},270^{\circ}\}$.
\end{comment}

\textit{Jigsaw} \cite{noroozi2016unsupervised, carlucci2019domain}: 
%\SL{Please write in the way that I wrote for Selfie.} 
By dividing an image into different patches, 
\textit{Jigsaw} trains a classifier to predict the correct permutation of  these patches.

\begin{comment}

\begin{align}\label{eq: jigsaw}
    \begin{array}{l}
\displaystyle\minimize_{\boldsymbol{\theta}} ~\mathbb E_{\mathbf x \in \mathcal D_\mathrm{p}} \left [\ell_{\mathrm{CE}}(J_{k\times k}(\boldsymbol{x}), p_j\in\mathcal{P};\boldsymbol{\theta}) \right ], 
    \end{array}
\end{align}
where $J_{k\times k}(\boldsymbol{x})$ is a process, using a $k\times k$ grid to decomposes the input image in $k^2$ patches which are random shuffled and used to form image with original size; $\ell_{\mathrm{CE}}(J_{k\times k}(\boldsymbol{x}), p_j\in\mathcal{P};\boldsymbol{\theta})$ is the cross-entropy between the permutation prediction and the ground-truth label $p_j\in\mathcal{P}$; $\mathcal{P}$ is the set of all patch permutations \footnote{It is a permutation of seqence $1,2\cdots,k^2$, as shown in \cite{carlucci2019domain}.}. (In our case, $|\mathcal{P}|=30$, $k=4$)

\end{comment}

\paragraph{Supervised Fine-tuning} %%\Def \{ (\mathbf x_i, y_i)\}_{i=1}^n
Let $\mathbf r (\mathbf x; \boldsymbol{\theta}_{\mathrm{p}})$ denote the mapping (parameterized by $\boldsymbol{\theta}_{\mathrm{p}}$) from an input sample $\mathbf x$ to its embedding space learnt from the self-supervised pretraining task $\mathcal T_{\mathrm{p}}$. 
Given    
a target \underline{f}inetuning task  $\mathcal T_{\mathrm{f}}$ with the
{labeled}   dataset $\mathcal D_{\mathrm{f}} $, the goal of fine-tuning  is to 
determine a classifier, parameterized by $\boldsymbol{\theta}_{\mathrm{f}}$, which maps the represetnation $\mathbf r (\mathbf x; \boldsymbol{\theta}_{\mathrm{p}})$ to the \textit{label} space. 
To learn the classifier, 
one can minimize a common supervised training loss  $\ell_{\mathrm{f}}(\boldsymbol{\theta}_{\mathrm{p}},\boldsymbol{\theta}_{\mathrm{f}}; \mathcal D_{\mathrm{f}})$   with a \textit{fixed} or \textit{re-trainable} model $\boldsymbol{\theta}_{\mathrm{p}}$, corresponding to \textit{partial fine-tuning} and \textit{full fine-tuning}, respectively.

\paragraph{AT versus standard training (ST)}
AT is known as one of the most powerful methods to train
 a robust classifier   against adversarial attacks \cite{madry2017towards,athalye2018obfuscated}. 
Considering an $\epsilon$-tolerant   $\ell_\infty$ attack $\boldsymbol{\delta}$ subject to $\| \boldsymbol{\delta} \|_{\infty} \leq \epsilon$, an adversarial example of a benign input $\mathbf x$ is   given by $\mathbf x + \boldsymbol{\delta}$. With the aid of adversarial examples, AT solves a min-max optimization problem of the generic form
%is cast as a min-max optimization problem, where the inner maximization problem generates adversarial perturbations,  and the outer minimization problem optimizes  an ML/DL model under the perturbed inputs.  That is,
\begin{align}\label{eq: adv_train}
    \begin{array}{l}
\displaystyle\minimize_{\boldsymbol{\theta}} ~\mathbb E_{\mathbf x \in \mathcal D} \left [   \maximize_{ \| \boldsymbol{\delta} \|_{\infty} \leq \epsilon }  ~  \ell(\boldsymbol{\theta}, \mathbf  x + \boldsymbol{\delta}) \right ], 
    \end{array}
\end{align}
where $\boldsymbol{\theta} $ denotes the   parameters of an ML/DL model, $\mathcal D$ is a given dataset, and $\ell$ signifies a classification loss evaluated at the model $\boldsymbol{\theta}$ and the perturbed input $\mathbf x + \boldsymbol{\delta}$. 
%\SL{Need to clarify the use of labeled and unlabeled data in AT.}
By fixing $\boldsymbol{\delta} = 0$, problem \eqref{eq: adv_train} then simplifies to the ST framework $\minimize_{\boldsymbol{\theta}} ~\mathbb E_{\mathbf x \in \mathcal D} \left [    \ell(\boldsymbol{\theta}, \mathbf  x) \right ]$.
\iffalse
It is  worth mentioning that   AT is not the only way to robustify a network, other approaches like randomized smoothing \cite{pmlr-v97-cohen19c} will also be considered under some circumstances in our work.
% \begin{align}\label{eq: st_train}
%     \begin{array}{l}
% \displaystyle\minimize_{\boldsymbol{\theta}} ~\mathbb E_{\mathbf x \in \mathcal D} \left [    \ell(\boldsymbol{\theta}, \mathbf  x) \right ], 
%     \end{array}
%     \tag{AT}
% \end{align}
% It is clear from \eqref{eq: adv_train} that AT can be applied to both {self-supervised pretraining} and {fine-tuning}.
We elaborate on all possible integrations of AT with {self-supervised pretraining} and fine-tuning in the next section.
\fi

%  \SL{Discuss the need of research on enforcing robustness in pretraining+fine-tuning, and the goal to explore the relationship between AT and pretraining+fine-tuning.}

\subsection{AT meets self-supervised pretraining and fine-tuning}

% AT  can be applied to either \textit{self-supervised pretraining} or \textit{supervised fine-tuning}. 

AT  given by \eqref{eq: adv_train} can be specified for either \textit{self-supervised pretraining} or \textit{supervised fine-tuning}. For example, AT for self-supervised pretraining
can be cast as problem \eqref{eq: adv_train}   by letting $\boldsymbol{\theta} \Def [ \boldsymbol{\theta}_{\mathrm{p}}^T, \boldsymbol{\theta}_{\mathrm{pc}}^T]^T$ and $\mathcal D \Def \mathcal{D}_{\mathrm{p}}$, and specifying $\ell$ as $\ell_{\mathrm{p}}$. 
In Table\,\ref{table: AT_self}, we summarize all the possible scenarios when AT meets self-supervised pretraining. 

\begin{table}[htb]
\centering
\caption{Summary of self-supervised pretraining scenarios. %{\red[Sahu et al uses small ``t"]}
}
\label{table: AT_self}
\begin{adjustbox}{max width=0.47\textwidth }
\begin{threeparttable}
\begin{tabular}{|c|c|c|c|c|}
\hline
Scenario & 
 \begin{tabular}[c]{@{}c@{}}Pretraining\\method \end{tabular}   & \begin{tabular}[c]{@{}c@{}} Loss $\ell$ in \eqref{eq: adv_train}
\end{tabular}       & \begin{tabular}[c]{@{}c@{}} Variables $\boldsymbol{\theta}$  in \eqref{eq: adv_train} \end{tabular} 
&     \begin{tabular}[c]{@{}c@{}}dataset\\$\mathcal D$ in \eqref{eq: adv_train}  \end{tabular} 
\\ \hline  
% ZO-GD  \cite{nesterov2015random}
% & nonconvex, unconstrained 
% &  {GauGE$^{1}$}
% &  $ O\left (\frac{1}{\sqrt{dT}} \right )$ 
% & $\mathbb E [ \| \nabla f(\mathbf x_T) \|_2^2] = O\left (\frac{d}{T} \right )$  & $O\left (T \right ) $ 
% %$O\left (|\mathcal B|  T \right ) $ 
% \\ \hline
$\mathcal P_1$ & None$^{1}$ & 
NA$^{2}$
& NA
& NA
\\ \hline
$\mathcal P_2$ & \begin{tabular}[c]{@{}c@{}} ST$^3$ \end{tabular}  & 
\begin{tabular}[c]{@{}c@{}} $\ell_{\mathrm{p}}$ \end{tabular} 
& $[ \boldsymbol{\theta}_{\mathrm{p}}^T, \boldsymbol{\theta}_{\mathrm{pc}}^T]^T$
& $\mathcal D_{\mathrm{p}}$
\\ \hline
$\mathcal P_3$ & \begin{tabular}[c]{@{}c@{}}  AT    \end{tabular}  & 
\begin{tabular}[c]{@{}c@{}} $\ell_{\mathrm{p}}$ \end{tabular} 
& $[ \boldsymbol{\theta}_{\mathrm{p}}^T, \boldsymbol{\theta}_{\mathrm{pc}}^T]^T$
& $\mathcal D_{\mathrm{p}}$
  \\ \hline 
\end{tabular}
\begin{tablenotes}
    \small
     \item[1] {\small None: the model form of $\boldsymbol{\theta}_{\mathrm{p}}$  is known in advance.}   
    \item[2] {\small NA: Not applicable.}
    \item[3] {\small ST: A special case of   \eqref{eq: adv_train} with $\boldsymbol{\delta} = \mathbf 0$.}
\end{tablenotes}
\end{threeparttable}
\end{adjustbox}
\vspace*{-0.1in}
\end{table}

\begin{table}[htb]
\centering
\caption{Summary of fine-tuning scenarios. %{\red[Sahu et al uses small ``t"]}
}
\label{table: AT_fine}
\begin{adjustbox}{max width=0.47\textwidth }
\begin{threeparttable}
\begin{tabular}{|c|c|c|c|c|c|}
\hline
Scenario &  \begin{tabular}[c]{@{}c@{}}Fine-tuning\\type \end{tabular}  &
 \begin{tabular}[c]{@{}c@{}}Fine-tuning\\method \end{tabular}         & \begin{tabular}[c]{@{}c@{}} Loss\\$\ell$ in \eqref{eq: adv_train}
\end{tabular}       & \begin{tabular}[c]{@{}c@{}} Variables\\$\boldsymbol{\theta}$  in \eqref{eq: adv_train} \end{tabular} 
&     \begin{tabular}[c]{@{}c@{}}dataset\\$\mathcal D$ in \eqref{eq: adv_train}  \end{tabular} 
\\ \hline  
% ZO-GD  \cite{nesterov2015random}
% & nonconvex, unconstrained 
% &  {GauGE$^{1}$}
% &  $ O\left (\frac{1}{\sqrt{dT}} \right )$ 
% & $\mathbb E [ \| \nabla f(\mathbf x_T) \|_2^2] = O\left (\frac{d}{T} \right )$  & $O\left (T \right ) $ 
% %$O\left (|\mathcal B|  T \right ) $ 
% \\ \hline
$\mathcal F_1$ & \begin{tabular}[c]{@{}c@{}} Partial\\ (with \textit{fixed} $\boldsymbol{\theta}_{\mathrm{p}}$)$^1$ \end{tabular} & 
ST & $\ell_{\mathrm{f}}$
& $\boldsymbol{\theta}_{\mathrm{f}}$
& $\mathcal{D}_{\mathrm{f}}$
\\ \hline
$\mathcal F_2$ & \begin{tabular}[c]{@{}c@{}} Partial\\ (with \textit{fixed} $\boldsymbol{\theta}_{\mathrm{p}}$) \end{tabular} & 
AT & $\ell_{\mathrm{f}}$
& $\boldsymbol{\theta}_{\mathrm{f}}$
& $\mathcal{D}_{\mathrm{f}}$
\\ \hline
$\mathcal F_3$ & \begin{tabular}[c]{@{}c@{}} Full$^2$\end{tabular} & 
ST & $\ell_{\mathrm{f}}$
& $[\boldsymbol{\theta}_{\mathrm{p}}^T, \boldsymbol{\theta}_{\mathrm{f}}^T ]^T$
& $\mathcal{D}_{\mathrm{f}}$
  \\ \hline 
  $\mathcal F_4$ & \begin{tabular}[c]{@{}c@{}} Full\end{tabular} & 
AT & $\ell_{\mathrm{f}}$
& $[\boldsymbol{\theta}_{\mathrm{p}}^T, \boldsymbol{\theta}_{\mathrm{f}}^T ]^T$
& $\mathcal{D}_{\mathrm{f}}$
  \\ \hline 
\end{tabular}
\begin{tablenotes}
   % \small
     \item[1] {\small \textit{Fixed} $\boldsymbol{\theta}_{\mathrm{p}}^*$ signifies the model learnt in a given pretraining scenario.}
     \item[2] {\small Full fine-tuning retrains $\boldsymbol{\theta}_{\mathrm{p}}$.}
\end{tablenotes}
\end{threeparttable}
\end{adjustbox}
\vspace*{-0.1in}
\end{table}

Given a pretrained model $\boldsymbol{\theta}_{\mathrm{p}}$, adversarial fine-tuning could have two forms:  a) AT for partial fine-tuning and b) AT for full fine-tuning. Here the former case a) solves a supervised fine-tuning task under the fixed model ($\boldsymbol{\theta}_{\mathrm{p}}$), and 
%learnt from the self-supervised pretraining stage. 
the latter case b) solves a supervised fine-tuning task by retraining $\boldsymbol{\theta}_{\mathrm{p}}$. 
In Table\,\ref{table: AT_fine}, we summarize different scenarios when AT meets supervised fine-tuning.

% given by
%  problem   \eqref{eq: adv_train} with $\boldsymbol{\theta} \Def   \boldsymbol{\theta}_{\mathrm{s}}$ and $\mathcal D \Def \mathcal{D}_{\mathrm{s}}$, and specifying $\ell$ as $\ell_{\mathrm{s}}$; b)  AT for full fine-tuning corresponds to  problem   \eqref{eq: adv_train} with $\boldsymbol{\theta} \Def [ \boldsymbol{\theta}_{\mathrm{p}}^T, \boldsymbol{\theta}_{\mathrm{s}}^T]^T$  and $\mathcal D \Def \mathcal{D}_{\mathrm{s}}$, and specifying $\ell$ as $\ell_{\mathrm{s}}$.

% It is clear from Table\,\ref{table: AT_self} and Table\,\ref{table: AT_fine} that every combination of a pretraining scenario ($\mathcal P_i$) and a fine-tuning scenario ($\mathcal F_j$) forms an integration of ST/AT with the self-supervised pretraining+fine-tuning scheme, where $i \in [3]$ and $j \in [4]$.   Here for ease of notation, let $[n]$ denote the integer set $\{1,2,\ldots, n \}$. 
% We highlight that \SL{[mention difference with Dawn Song's work. Their work dose not enjoy the proerty that pretraining can be separated from fine-tuning. Instead they consider  AT for jointly optimizing pretraining + fine-tuning losses with respect to a supervised classification task only.]}
It is worth noting that our study on the integration of AT with a pretraining+fine-tuning scheme  $ (\mathcal P_i, \mathcal F_j) $ provided by Tables\,\ref{table: AT_self}-\ref{table: AT_fine} is different from  % the  integration of  AT with the self-supervised  task (specified for \textit{rotation}) in 
\cite{hendrycks2019using}, which
%The previous work  \cite{hendrycks2019using} 
conducted  one-shot AT over a   supervised classification task integrated with 
a rotation  self-supervision  task.

% regularized with  a self-supervised  training loss
% augmenting the the standard classification loss
% with the self-supervised  training loss, only corresponding to  the   target supervised task.
% Thus, it did not enjoy the benefit of the pretraining+fine-tuning scheme,  which separates a pretraininig stage (that allows to use unlabeled data) from a target supervised classification task.
%co-optimizes the classification loss and the self-supervised  training loss in 
% does not enjoy the beneficial properties that pretraining can be separated from fine-tuning. 

%Instead they consider AT for jointly optimizing classification loss with a separate auxiliary self-supervised rotation (\ref{eq: rotation}) loss together. Although it takes a little advantage on robust accuracy, our approach surpasses it consistently in terms of standard accuracy and other $19$ unforeseen attacks in \textit{Rotation} and \textit{Jigsaw} tasks, as shown in Figure~\ref{fig:rot_unforeseen} and.~\ref{fig:jig_unforeseen}.
 
In order to explore the network robustness against different configurations $\{ (\mathcal P_i, \mathcal F_j) \}$,
\textbf{we ask}: 
%\TL{We have mention Q1 and Q3 on the introduction, but not for Q2.}
\textit{is AT for robust pretraining sufficient to boost the adversarial robustness of fine-tuning?}
\textit{What is the influence of   fine-tuning strategies (partial or full) on the adversarial robustness of image classification?}
\textit{How does the type   of self-supervised pretraining task  affect the classifier's robustness?}

We provide detailed answers to the above questions in Sec.\,\ref{sec: answer1}, Sec.\,\ref{sec: answer2} and Sec.\,\ref{sec: answer3}. In a nutshell, we find that robust representation learnt from adversarial pretraining is transferable to down-stream fine-tuning tasks to some extent. However, a more significant robustness improvement is obtained by adversarial fine-tuning. Moreover, AT for full fine-tuning outperforms that for partial fine-tuning in terms of both robust accuracy and standard accuracy (except the \textit{Jigsaw}-specified self-supervision task). Furthermore, different self-supervised tasks demonstrate diverse adversarial vulnerability. As will be evident later, such diversified tasks provide complementary benefits to model robustness and therefore can be combined.
   
\begin{figure*}[t]
\begin{center}
   \includegraphics[width=0.85\linewidth]{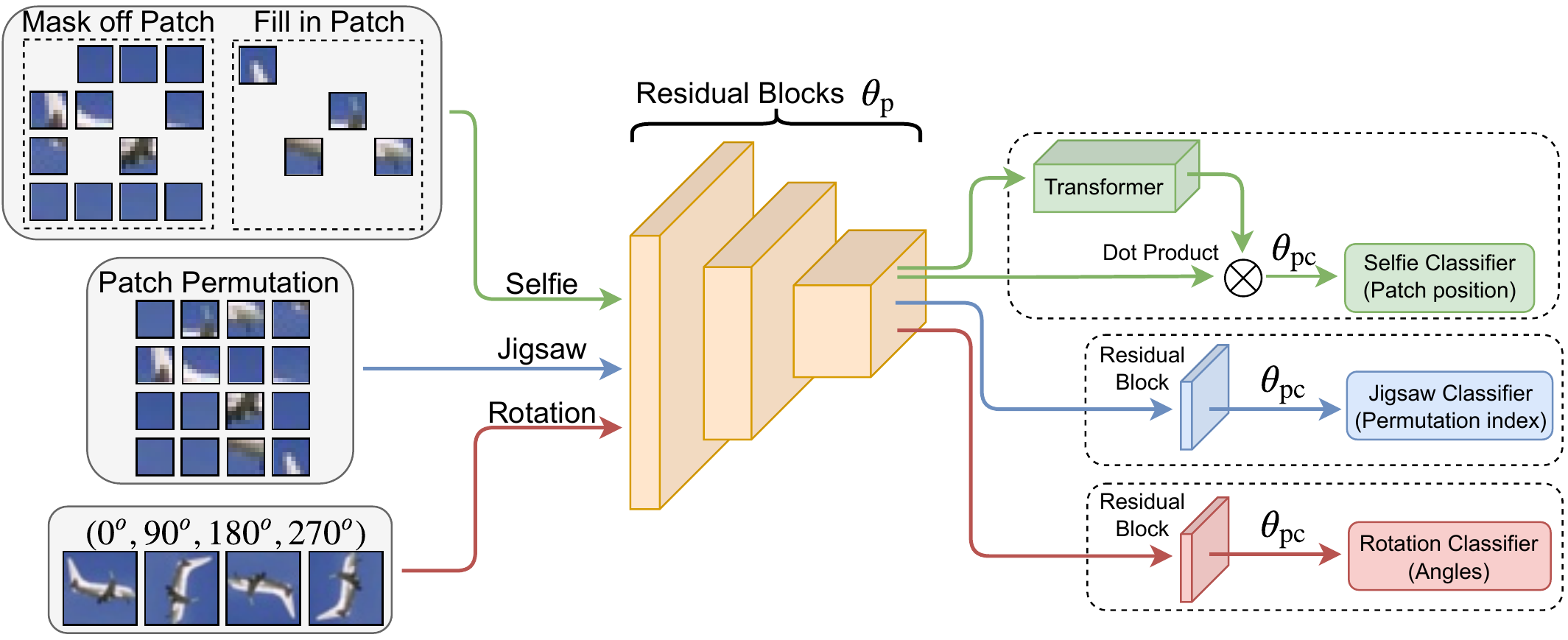}
\end{center}
\vspace{-3mm}
   \caption{The overall framework of ensemble adversarial pretraining. The pretrained weights $\theta_\mathrm{p}$ are the first three blocks of ResNet-50v2 \cite{he2016deep}; \textit{Green arrows} (\textcolor{green}{\ding{212}}), \textit{Blue arrows} (\textcolor{blue}{\ding{212}}) and \textit{Red arrows} (\textcolor{red}{\ding{212}}) represent the feed forward paths of \textit{Selfie}, \textit{Jigsaw} and \textit{Rotation}, respectively.}
\label{fig:ensemble}
\vspace{-3mm}
\end{figure*}

\subsection{AT by leveraging ensemble of multiple self-supervised learning tasks}

In what follows, we generalize AT  to learn a robust pretrained model by leveraging the diversified pretraining tasks. More specifically, consider $M$ self-supervised pretraining tasks $\{ \mathcal T_{\mathrm{p}}^{(i)} \}_{i=1}^M$, each of which obeys the formulation in Section\,\ref{sec: set_up}. We  generalize problem \eqref{eq: adv_train} to
\begin{align}\label{eq: prob_multiple_self}
    \begin{array}{l}
        \displaystyle\minimize_{\boldsymbol{\theta}_r, \{ \boldsymbol{\theta}_{\mathrm{pc}}^{(i)} \}}   ~  \mathbb E_{\mathbf x \sim \mathcal D_p} \left [ 
        %\maximize_{\{   \|  \boldsymbol{\delta}^{(i)} \|_\infty \leq \epsilon\}_{i=1}^M }
        ~ \mathcal{L}_{\mathrm{adv}}( \boldsymbol{\theta}_{\mathrm{p}}, \{ \boldsymbol{\theta}_{\mathrm{pc}}^{(i)} \}, \mathbf x)
        \right ], % , \{ \mathbf x + \boldsymbol{\delta}^{(i)} \}_{i=1}^M
        % \\  \left (  \sum_{i=1}^M  w_i f_{\mathrm{p}}^{(i)} (\boldsymbol{\theta}_{\mathrm{p}}, \boldsymbol{\theta}_{\mathrm{pc}}^{(i)}; \mathbf x + \boldsymbol{\delta}^{(i)} ) + \lambda h( \boldsymbol{\theta}_{\mathrm{r}}, \{ \boldsymbol{\theta}_o^{(i)} \}_i, \{ \boldsymbol{\delta}^{(i)} \}_i ) \right )  ,
     \end{array}
\end{align}
where $\mathcal{L}_{\mathrm{adv}}$ denotes the adversarial loss given by
%\SL{Do not use $\mathcal D$ to denote diversity term since we reserve it for dataset.}
\begin{align}\label{eq: loss_multiple_self}
   & \mathcal{L}_{\mathrm{adv}}( \boldsymbol{\theta}_{\mathrm{p}}, \{ \boldsymbol{\theta}_{\mathrm{pc}}^{(i)} \}, \mathbf x) \nonumber \\
    \Def &
    \maximize_{\{   \|  \boldsymbol{\delta}^{(i)} \|_\infty \leq \epsilon\}} \sum_{i=1}^M   \ell_{\mathrm{p}}^{(i)} (\boldsymbol{\theta}_{\mathrm{p}}, \boldsymbol{\theta}_{\mathrm{pc}}^{(i)}, \mathbf x + \boldsymbol{\delta}^{(i)} ) \nonumber \\
    & \hspace*{0.8in} + \lambda  g( \boldsymbol{\theta}_{\mathrm{p}}, \{ \boldsymbol{\theta}_{\mathrm{pc}}^{(i)} \}, \{ \boldsymbol{\delta}^{(i)} \} ).
\end{align}
In \eqref{eq: prob_multiple_self},
for ease of notation, we replace $\{ \cdot\}_{i=1}^M$ with $\{ \cdot\}$,  $\boldsymbol{\theta}_{\mathrm{p}}$
denotes the common network shared among different self-supervised tasks, and $\boldsymbol{\theta}_{\mathrm{pc}}^{(i)}$ denotes a sub-network customized for the $i$th  task. We refer readers to Figure~\ref{fig:ensemble} for an overview of our proposed model architecture. 
In \eqref{eq: loss_multiple_self}, $ \ell_{\mathrm{p}}^{(i)}$ denotes the $i$th pretraining loss, $g$ denotes a \textit{diversity-promoting regularizer}, and $\lambda \geq 0$ is a regularization parameter.  Note that $\lambda = 0$ gives the averaging ensemble strategy. In our case, we perform grid search to tune $\lambda$ around the value chosen in \cite{pang2019improving}. Details are referred to the supplement. 
% \SL{[Better to cite which section in supp.]}. 
  
Spurred by \cite{pang2019improving,wang2019unified}, we  quantify the diversity-promoting regularizer $g$
through the orthogonality of input gradients of different self-supervised pretraining losses,
\begin{align}\label{eq: OPDM}
 \begin{array}{l}
         g( \boldsymbol{\theta}_{\mathrm{p}}, \{ \boldsymbol{\theta}_{\mathrm{pc}}^{(i)} \}, \{ \boldsymbol{\delta}^{(i)} \} ) \Def \log \mathrm{det}( \mathbf G^T \mathbf G  ),
 \end{array}
\end{align}
where each column of $\mathbf G $  corresponds to a \textit{normalized} input gradient $\{ \nabla_{\boldsymbol{\delta}_i} \ell_{p}^{(i)}(\boldsymbol{\theta}_{\mathrm{p}}, \boldsymbol{\theta}_{\mathrm{pc}}^{(i)}, \mathbf x + \boldsymbol{\delta}^{(i)} ) \}$, and  $ g$ reaches the maximum value $0$ as input gradients become orthogonal, otherwise it is negative.
%   We measure the diversity between adversarial attacks through the similarity between perturbation directions, namely, input gradients $\{ \nabla_{\boldsymbol{\delta}_i} f_{p}^{(i)}(\boldsymbol{\theta}_{\mathrm{p}}, \boldsymbol{\theta}_{\mathrm{pc}}^{(i)}, \mathbf x + \boldsymbol{\delta}^{(i)} ) \}$. 
The rationale behind the diversity-promoting adversarial loss \eqref{eq: loss_multiple_self} is that we aim to design a robust model $\boldsymbol{\theta}_{\mathrm{p}}$ by defending attacks from diversified perturbation directions.

\section{Experiments and Results}
In this section, we design and conduct extensive experiments to examine the network robustness against different configurations $\{ (\mathcal{P}_i,\mathcal{F}_j) \}$ for image classification. %datasets  CIFAR-10 \cite{krizhevsky2009learning} and ImageNet \cite{russakovsky2015imagenet}. 
\textit{First}, we show adversarial self-supervised pretraining (namely, $\mathcal{P}_3$ in Table\,\ref{table: AT_self}) improves the performance of downstream tasks. We also discuss the influence of different fine-tuning strategies $\mathcal{F}_j$ on the adversarial robustness. \textit{Second}, we show the diverse impacts of different self-supervised tasks on their resulting pretrained models. \textit{Third}, we ensemble those self-supervised tasks to perform adversarial pretraining. At the fine-tuning phase, we also ensemble three best models with the configuration ($\mathcal{P}_3$, $\mathcal{F}_4$) and show its performance superiority. \textit{Last}, we report extensive ablation studies to reveal the influence of the size of the datasets $\mathcal{D_{\mathrm{p}}}$ and the resolution of images in $\mathcal{D_{\mathrm{p}}}$, as well as other defense options beyond AT.

\subsection{Datasets}
\textbf{Dataset Details} We consider four different datasets in our experiments: CIFAR-10, CIFAR-10-C \cite{hendrycks2019robustness}, CIFAR-100 and \textbf{R-ImageNet-224} (a specifically constructed ``restricted" version of ImageNet, with resolution $224 \times 224$). For the last one, we indeed to demonstrate our approach on high-resolution data despite the computational challenge. We follow \cite{santurkar2019computer} to choose 10 super classes which contain a total of 190 ImageNet classes. The detailed classes distribution of each super class can be found in our supplement.
% \textcolor{red}{\cite{liu2019towards} shows that ``\textit{The feasibility of transferability is related to the similarity of both input and label}". Thus, we use ``Ship", ``Truck", ``Airplane" and ``Automobile" super-classes to replace the ``Turtle", ``Primate" and ``Crab" super-classes in \cite{santurkar2019computer}'s setting. For each super-class, we will randomly sample 3,000 images for training and 1,000 for testing. The detailed classes distribution of each super class can be found in our supplement.} 

For the ablation study of different pretraining dataset sizes, we sample more training images from the 80 Million Tiny Images dataset \cite{torralba200880} where CIFAR-10 was selected from. Using the same 10 super classes, we form CIFAR-30K (\ie, 30,000 for images), CIFAR-50K, CIFAR-150K for training, and keep another 10,000 images for hold-out testing.

\textbf{Dataset Usage} 
In Sec.\,\ref{sec: answer1}, Sec.\,\ref{sec: answer2} and Sec.\,\ref{sec: answer3}, for all results, we use CIFAR-10 training set for both pretraining and fine-tuning. We evaluate our models on the CIFAR-10 testing set and CIFAR-10-C. In Sec.\,\ref{sec: abla}, we use CIFAR-10, CIFAR-30K, CIFAR-50K, CIFAR-150K and R-ImageNet-224 for pretraining, and CIFAR-10 training set for fine-tuning, while evaluating on CIFAR-10 testing set. We also validate our approaches on CIFAR-100 in the supplement. In all of our experiments, we randomly split the original training set into a training set and a validation set (the ratio is 9:1).

\subsection{Implementation Details}
\textbf{Model Architecture:} For pretraining with the \textit{Selfie} task, we identically follow the setting in \cite{trinh2019selfie}. For \textit{Rotation} and \textit{Jigsaw} pretraining tasks, we use ResNet-50v2 \cite{he2016identity}. %\underline{ii)} Fine-tuning Network Architecture: 
For the fine-tuning, we use ResNet-50v2 for all. Each fine-tuning network will inherit the corresponding robust pretrained weights to initialize the first three blocks of ResNet-50v2, while leaving the remaining blocks randomly initialized.

%The different fine-tuning network will inherit robust weights of the first three blocks of ResNet-50v2 from its own pretraining tasks and have a random initialized the fourth block of ResNet-50v2.

\textbf{Training \& Evaluation Details:} All pretraining and fine-tuning tasks are trained using SGD with 0.9 momentum. We use batch sizes of 256 for CIFAR-10, ImageNet-32 and 64 for R-ImageNet-224.
All pretraining tasks adopt cosine learning rates. The maximum and minimum learning rates are 0.1 and $10^{-6}$ for \textit{Rotation} and \textit{Jigsaw} pretraining; 0.025 and $10^{-6}$ for \textit{Selfie} pretraining; and 0.001 and $10^{-8}$ for ensemble pretraining. All fine-tuning phases follow a multi-step learning rate schedule, starting from 0.1 and decayed by 10 times at epochs 30 and 50 for a 100 epochs training.

%For \textit{Rotation} and \textit{Jigsaw} pretraining, the maximum and minimum learning rates are 0.1 and $10^{-6}$ each. For \textit{Selfie} pretraining, they are 0.025 and maximum learning rate is 0.025 and the minimum learning rate is $10^{-6}$. For ensemble pretraining, the maximum learning rate is 0.001 and the minimum learning rate is $10^{-8}$. 
%All fine-tuning tasks multi-step learning rate decay from 0.1. The decay happens at 30, 50 epochs and the learning rate after decay is a tenth of the previous learning rate. We tune all hyperparamaters using the grid search method. 
%\underline{iv)} Adversarial Attack: 

We use 10-step and 20-step $\ell_\infty$ PGD attacks \cite{madry2017towards} for adversarial training and evaluation, respectively. Unless otherwise specified, we follow \cite{hendrycks2019using}'s setting with $\epsilon=\frac{8.0}{225}$ and $\alpha=\frac{2.0}{255}$. For all adversarial evaluations, we use the full testing datasets ($i.e$ $10,000$ images for CIFAR-10) to generate adversarial images. We also consider unforeseen attacks \cite{kang2019testing,hendrycks2019robustness}.

\textbf{Evaluation Metrics \& Model Picking Criteria:} We follow \cite{zhang2019theoretically} to use: i) \textit{Standard Testing Accuracy (TA)}: the classification accuracy on the clean test dataset; II) \textit{Robust Testing Accuracy (RA)}: the classification accuracy on the attacked test dataset. In our experiments, we use TA to pick models for a better trade-off with RA. Results of models picked using RA criterion are included in the supplement.

%\subsection{Self-supervised Pretraining Helps}
\subsection{{Adversarial self-supervised pertraining \& fine-tuning   helps classification robustness} } \label{sec: answer1}

% \TL{[Question: Shall we compare with jointly optimizing method in terms of epochs? i.e. rotation jointly optimizing need 63 epochs to achieve the best RA, which is more than scenario ($\mathcal{P}_3,\mathcal{F}_4$)].} 

% \SL{SL: What dataset in Table\,\ref{table:p3_matter}, do not forget to mention it.}

\begin{table*}[t]
\begin{center}
\caption{Evaluation Results of Eight Different ($\mathcal{P}_i,\mathcal{F}_j$) Scenarios. Table~\ref{table: AT_self} and Table~\ref{table: AT_fine} provide detailed definitions for $\mathcal{P}_1$ (without pre-training), $\mathcal{P}_2$ (standard self-supervision pre-training), $\mathcal{P}_3$ (adversarial self-supervision pre-training), $\mathcal{F}_1$ (partial standard fine-tuning), $\mathcal{F}_2$ (partial adversarial fine-tuning), $\mathcal{F}_3$ (full standard fine-tuning), and $\mathcal{F}_4$ (full adversarial fine-tuning). The best results are highlighted (\textcolor{red}{$1^{\mathrm{st}}$},\textcolor{blue}{$2^{\mathrm{nd}}$}) under each column of different self-supervised pretraining tasks.} 
% \vspace{-2.5mm}
\label{table:p3_matter}
\begin{threeparttable}
\resizebox{0.9\textwidth}{!}{
\begin{tabular}{c|c|c|c|c|c|c|c|c|c}
\hline
\multirow{2}{*}{Scenario} & \multicolumn{3}{c|}{\textit{Selfie} Pretraining} & \multicolumn{3}{c|}{\textit{Rotation} Pretraining} & \multicolumn{3}{c}{\textit{Jigsaw} Pretraining} \\ \cline{2-10} 
 & TA (\%) & RA (\%) & Epochs  & TA (\%) & RA (\%) & Epochs  & TA (\%) & RA (\%) & Epochs \\ \hline
($\mathcal{P}_1,\mathcal{F}_3$) & 94.24 & 0.00 & 92 & 94.24 & 0.00 & 92 & \textcolor{red}{94.24} & 0.00 & 92\\
($\mathcal{P}_1,\mathcal{F}_4$) & 84.72 & 47.22 & 99 & 84.72 & 47.22 & 99 & 84.72 & 47.22 & 99\\
\hline
% ($\mathcal{P}_2,\mathcal{F}_1$) & 92.43 & 0.00 & 91 & 92.88 & 0.00 & 93 & 84.11 & 0.00 & 92\\
% ($\mathcal{P}_2,\mathcal{F}_2$) & \textcolor{blue}{10.00} & \textcolor{blue}{10.00} & - & 54.53 & 19.29 & 91 & 27.36 & 12.05 & 86\\
($\mathcal{P}_2,\mathcal{F}_3$) & \textcolor{red}{95.09} & 0.00 & 97 & \textcolor{red}{95.45} & 0.00 & 92 & \textcolor{blue}{93.93} & 0.00 & 89\\
($\mathcal{P}_2,\mathcal{F}_4$) & 85.56 & \textcolor{blue}{50.42} & \textcolor{blue}{60} & 86.66 & \textcolor{red}{50.95} & \textcolor{red}{45} & 85.18 & \textcolor{red}{50.94} & \textcolor{red}{46}\\
\hline
($\mathcal{P}_3,\mathcal{F}_1$) & 78.93 & 6.30 & 82 & 86.83 & 18.22 & 99 & 80.47 & 2.68 & 87\\
($\mathcal{P}_3,\mathcal{F}_2$) & 74.30 & 37.65 & 64 & 82.32 & 45.10 & 47 & 72.76 & 32.59 & 51\\
($\mathcal{P}_3,\mathcal{F}_3$) &  \textcolor{blue}{94.69} & 0.00 & 86 & \textcolor{blue}{94.79} & 0.00 & 92 & 93.06 & 0.00 & 93\\
($\mathcal{P}_3,\mathcal{F}_4$) & 86.02 & \textcolor{red}{51.05} & \textcolor{red}{50} & 85.66 & \textcolor{blue}{50.40} & \textcolor{blue}{46} & 84.50 & \textcolor{blue}{49.61} & \textcolor{blue}{48}\\
\hline
\hline
\end{tabular}}
\end{threeparttable}
\end{center}
\vspace{-3mm}
\end{table*}

We systematically study all possible configurations of  pretraining and fine-tuning considered in Table\,\ref{table: AT_self} and Table\,\ref{table: AT_fine}, where recall that 
the expression $(\mathcal P_i, \mathcal F_j)$ denotes  a specified pretraining+fine-tuning scheme. The \textit{baseline} schemes are given by the end-to-end standard training (ST), namely, $(\mathcal P_1, \mathcal F_3)$ and the end-to-end adversarial training (AT), namely, $(\mathcal P_1, \mathcal F_4)$. Table\,\ref{table:p3_matter} shows TA, RA, and iteration complexity of fine-tuning (in terms of number of epochs) under different  pretraining+fine-tuning strategies involving different self-supervised pretraining tasks, \textit{Selfie}, \textit{Rotation} and \textit{Jigsaw}. In what follows, we analyze the results of Table\,\ref{table:p3_matter} and provide additional insights.

% \textit{Self-supervised pretraining improves RA and saves computation cost for final down-stream classification tasks.}
We begin by focusing on the scenario of integrating the \textit{standard pretraining} strategy $\mathcal P_2$ with  fine-tuning schemes $ \mathcal F_3$ and $\mathcal F_4$ used in baseline methods. Several observations can be made from the comparison $(\mathcal P_2, \mathcal F_3)$ vs. $(\mathcal P_1, \mathcal F_3)$  and  $(\mathcal P_2, \mathcal F_4)$ vs. $(\mathcal P_1, \mathcal F_4)$ in Table\,\ref{table:p3_matter}. 1) The use of self-supervised pretraining   consistently improves TA and/or RA even if only standard pretraining is conducted; 2) The use of adversarial fine-tuning $ \mathcal F_4$ (against standard fine-tuning $\mathcal F_3$) is crucial, leading to significantly improved RA under both $\mathcal P_1$ and $\mathcal P_2$; 3) Compared $(\mathcal P_1, \mathcal F_4)$  with  $(\mathcal P_2, \mathcal F_4)$, the use of self-supervised pretraining offers    better eventual model robustness   (around $3\%$ improvement)  and faster fine-tuning speed (almost saving the half number of epochs).

Next, we investigate how the \textit{adversarial pretraining} (namely, $\mathcal P_3$) affects the eventual model robustness.  
It is shown by  $(\mathcal P_3, \mathcal F_1)$ and $(\mathcal P_3, \mathcal F_2)$ in  Table\,\ref{table:p3_matter} that the robust feature representation learnt from $\mathcal P_3$ benefits adversarial robustness even in the case of {partial fine-tuning}, but the use of \textit{adversarial  partial fine-tuning}, namely, $(\mathcal P_3, \mathcal F_2)$, yields a $30\%$ more improvement. We also observe from the case of $(\mathcal P_3, \mathcal F_3)$  that the \textit{standard full fine-tuning} harms the robust feature representation learnt from $\mathcal P_3$, leading to $0\%$ RA. Furthermore, when the  \textit{adversairal full fine-tuning} is adopted, namely, $(\mathcal P_3, \mathcal F_4)$, the most significant robustness improvement is acquired. This observation is consistent with $(\mathcal P_2, \mathcal F_4)$  against  $(\mathcal P_2, \mathcal F_3)$.
  
Third, at the first glance, \textit{adversarial full fine-tuning} (namely, $\mathcal F_4$) is the most important step to improve the final mode   robustness. However, \textit{adversarial pretraining} is also a key,  particularly for reducing the computation cost   of fine-tuning; for example, less than $50$ epochs in $(\mathcal P_3, \mathcal F_4)$ vs. $99$ epochs in the end-to-end AT $(\mathcal P_1, \mathcal F_4)$.
  
Last but not the least, we note that the aforementioned results are consistent against different  self-supervised prediction  tasks. However, \textit{Selfie} and \textit{Rotation} are more favored than \textit{Jigsaw} to improve the final model robustness. For example, in the cases of adversarial pretraining followed by standard and adversarial partial fine-tuning, namely, $(\mathcal P_3, \mathcal F_1)$ and $(\mathcal P_3, \mathcal F_2)$,   \textit{Selfie} and \textit{Rotation} yields at least $3.5\%$ improvement in RA. As the adversarial full fine-tuning is used, namely, $(\mathcal P_3, \mathcal F_4)$,  \textit{Selfie} and \textit{Rotation} outperform  \textit{Jigsaw} in both TA and RA, where  \textit{Selfie} yields the largest improvement, around $2.5\%$ in both TA and RA.

\subsection{Comparison with one-shot AT regularized by self-supervised prediction task} \label{sec: answer2}

In what follows, we compare our proposed adversarial pretraining followed by adversarial fine-tuning approach, namely, $(\mathcal{P}_3$, $\mathcal{F}_4)$ in Table\,\ref{table:p3_matter}
with the one-shot AT 
that optimizes  a classification task regularized by   the self-supervised \textit{rotation} prediction task  \cite{hendrycks2019using}. In addition to evaluating this comparison in   TA and RA (evaluated at $\ell_\infty$ PGD attack \cite{madry2017towards}), we also measure the robustness in eventual classification against $12$ unforeseen attacks that are not used in AT \cite{kang2019testing}. More results can be found in the supplement.

\begin{figure}[ht]
\begin{center}
   \includegraphics[width=1\linewidth]{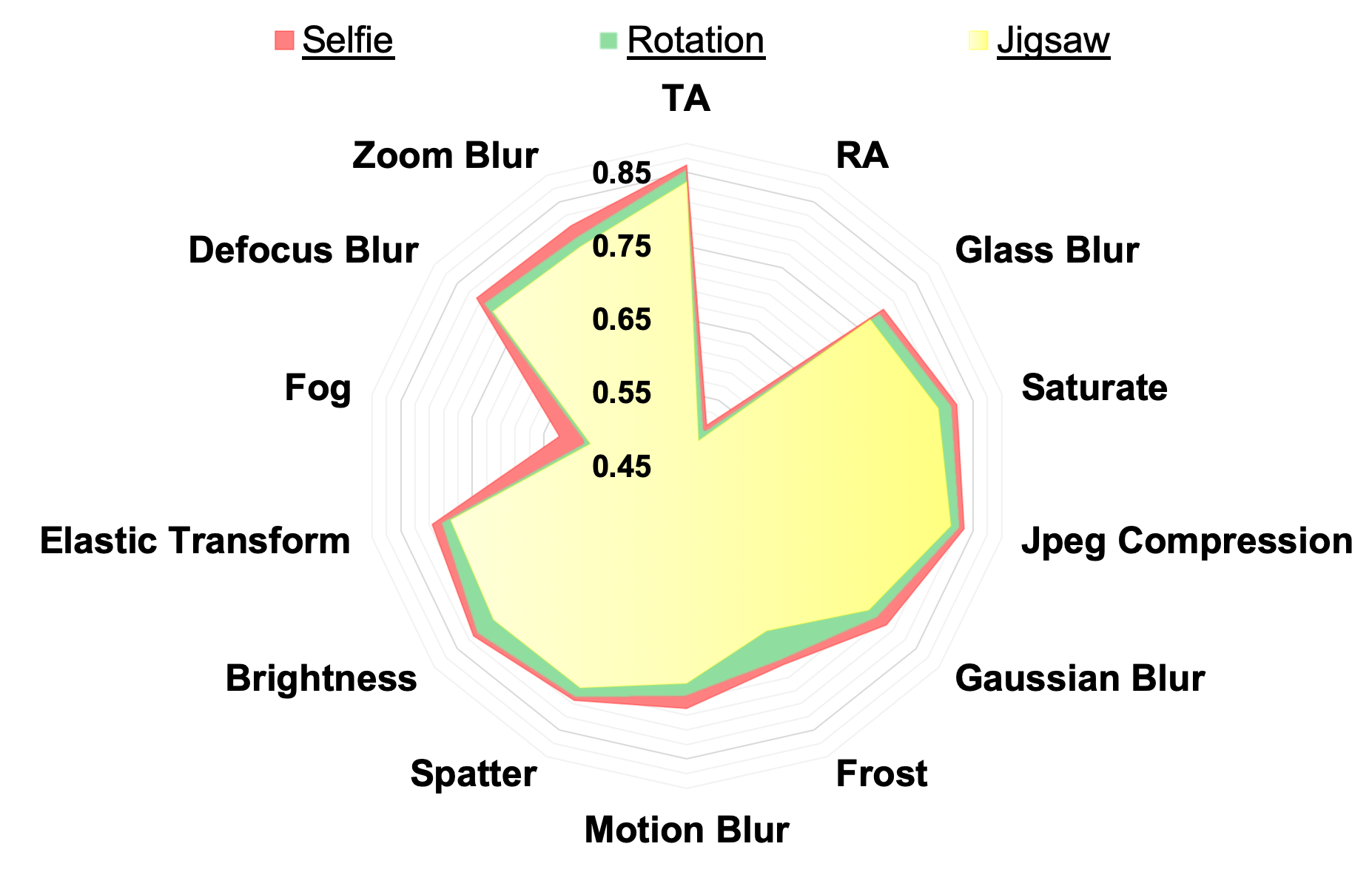}
\end{center}
   \caption{The summary of the accuracy over unforeseen adversarial attackers. Competition among adversarial fine-tuned models with \textit{Selfie}, \textit{Rotation} and \textit{Jigsaw} adversarial pretraining.}
\label{fig:unforeseen}
\vspace{-1mm}
\end{figure}

Figure~\ref{fig:rot_unforeseen}  presents the multi-dimensional performance comparison of our approach vs. the baseline method in \cite{hendrycks2019using}. As we can see, our approach yields $1.97\%$ improvement on TA while $0.74\%$ degradation on RA.  However, our approach yields consistent robustness improvement in defending all $12$  unforeseen attacks, where the improvement ranges from  $1.03\%$ to $6.53\%$. Moreover, our approach separates pretraining and fine-tuning such that the target image classifier can be learnt from a warm start, namely, the adversarial pretrained representation network. This mitigates the computation drawback of one-shot AT in \cite{hendrycks2019using},  recalling that our  advantage in saving computation cost was shown in Table\,\ref{table:p3_matter}. Next, Figure\,\ref{fig:unforeseen} presents the performance of our approach under different types of self-supervised prediction task. As we can see, \textit{Selfie} provides consistently better performance than others, where \textit{Jigsaw} performs the worst. 

%To further support the effectiveness of our approach against  unforeseen attacks, we conduct additional experiments by replacing  the \textit{rotation} task with the \textit{Jigsaw} task

%In \SL{Appendix\,xxxx}, 

% is the only work so farthat utilizes unlabeled data viaself supervisionto train arobust model given a target supervised classification task.It improves AT by leveraging the rotation prediction self-supervised as an auxiliary task, which is co-optimized withthe conventional AT loss.  Our self-supervised pretrainingand fine-tuning differ from all above settings

\begin{comment}
\textbf{Pretraining + Fine-tuning v.s. Jointly Training} As show in Figure~\ref{fig:rot_unforeseen} and Figure~\ref{fig:jig_unforeseen}, comparing with jointly optimizing, rotation ($\mathcal{P}_3$, $\mathcal{F}_4$) and Jigsaw ($\mathcal{P}_3$, $\mathcal{F}_4$) are 0.74 point and 2.87 points smaller in terms of RA, respectively. However, for TA and other 12 unforeseen adversarial attacks, ($\mathcal{P}_3$, $\mathcal{F}_4$) is consistently surpass jointly optimizing with a large marginal.
\end{comment}

\begin{figure}[ht]
\begin{center}
   \includegraphics[width=1\linewidth]{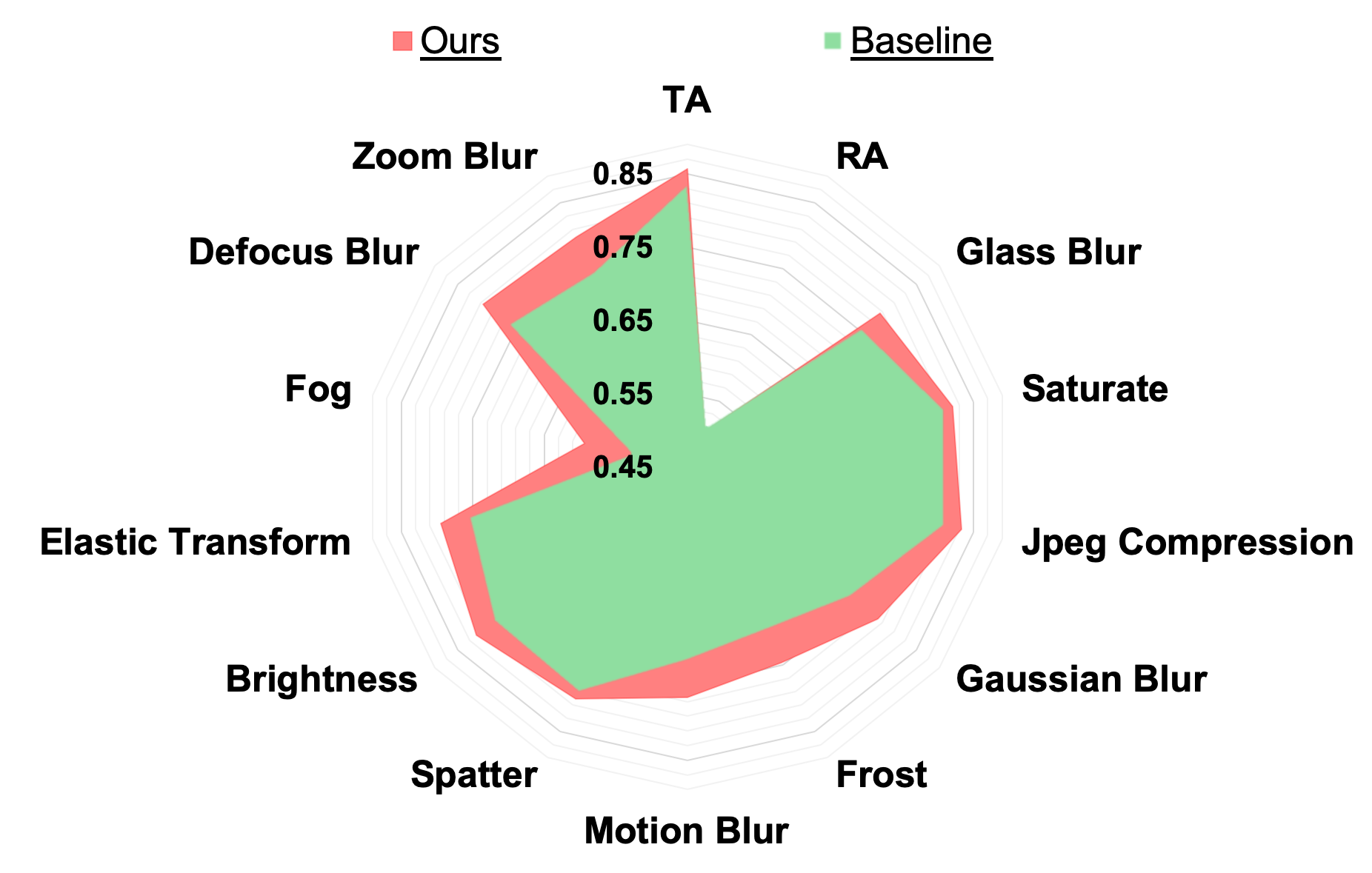}
\end{center}
   \caption{The summary of the accuracy over unforeseen adversarial attackers. Our models are obtained after adversarial fine-tuning with adversarial \textit{Rotation} pretraining. Baseline are co-optimized models with \textit{Rotation} auxiliary task \cite{hendrycks2019using}.}
\label{fig:rot_unforeseen}
\vspace{-3mm}
\end{figure}

\begin{comment}
\begin{figure}[ht]
\begin{center}
   \includegraphics[width=1\linewidth]{Pics/J_PvsCO.png}
\end{center}
   \caption{The summary of the accuracy over unforeseen adversarial attackers. Our models are obtained after adversarial fine-tuning with adversarial \textit{Jigsaw} pretraining. Baseline are co-optimized models with \textit{Jigsaw} auxiliary task \cite{hendrycks2019using}.}
\label{fig:jig_unforeseen}
\vspace{-3mm}
\end{figure}
\end{comment}

\subsection{Diversity vs. Task Ensemble} \label{sec: answer3}

In what follows, we show that different self-supervised prediction tasks demonstrate a \textit{diverse adversarial vulnerability} even if their corresponding RAs  remain similar. We evaluate such a diversity  through the transferability of adversarial examples generated from  robust classifiers fine-tuned from the adversarially pretrained models using different self-supervised prediction tasks. We then demonstrate the performance of our proposed adversarial pretraining method \eqref{eq: prob_multiple_self} by leveraging an ensemble  of  \textit{Selfie}, \textit{Rotation}, and \textit{Jigsaw}.
 
In Table\,\ref{table:diversity}, we present the transferbility of PGD attacks generated from  the final model trained using adversarial pretraining followed by adversarial  full fine-tuning, namely,
%$(\mathcal P_3, \mathcal F_2)$ and 
$(\mathcal P_3, \mathcal F_4)$, where for ease of presentation,  let $\mathrm{Model}(t)$ denote the classifier learnt using the self-supervised pretraining task $t \in \{ \text{\textit{Selfie}}, \text{\textit{rotation}} , \text{\textit{Jigsaw}} \}$. Given the PGD attacks from $\mathrm{Model}(t)$, we evaluate their transferbility, in terms of attack success rate (ASR\footnote{ASR is given by the ratio of \textit{successful} adversarial examples over the total number of $10,000$ test images.}), against $\mathrm{Model}(t^\prime)$. If $t^\prime = t$, then ASR reduces to $1-\mathrm{RA}$. If $t^\prime \neq  t$, then ASR reflects the attack transferbility from $\mathrm{Model}(t)$ to $\mathrm{Model}(t^\prime)$. As we can see, the diagonal entries of Table\,\ref{table:diversity} correspond to the largest ASR at each column. This is not surprising, since transferring to another model makes the attack being weaker. One interesting observation is that ASR suffers a larger drop when transferring attacks from $\mathrm{Model}(\textit{Jigsaw})$ to other target models. This implies that $\mathrm{Model}(\textit{Selfie})$ and $\mathrm{Model}(\textit{Rotation})$ yields better robustness, consistent with our previous results like Figure\,\ref{fig:unforeseen}.

\begin{figure}[ht]
\begin{center}
   \includegraphics[width=1\linewidth]{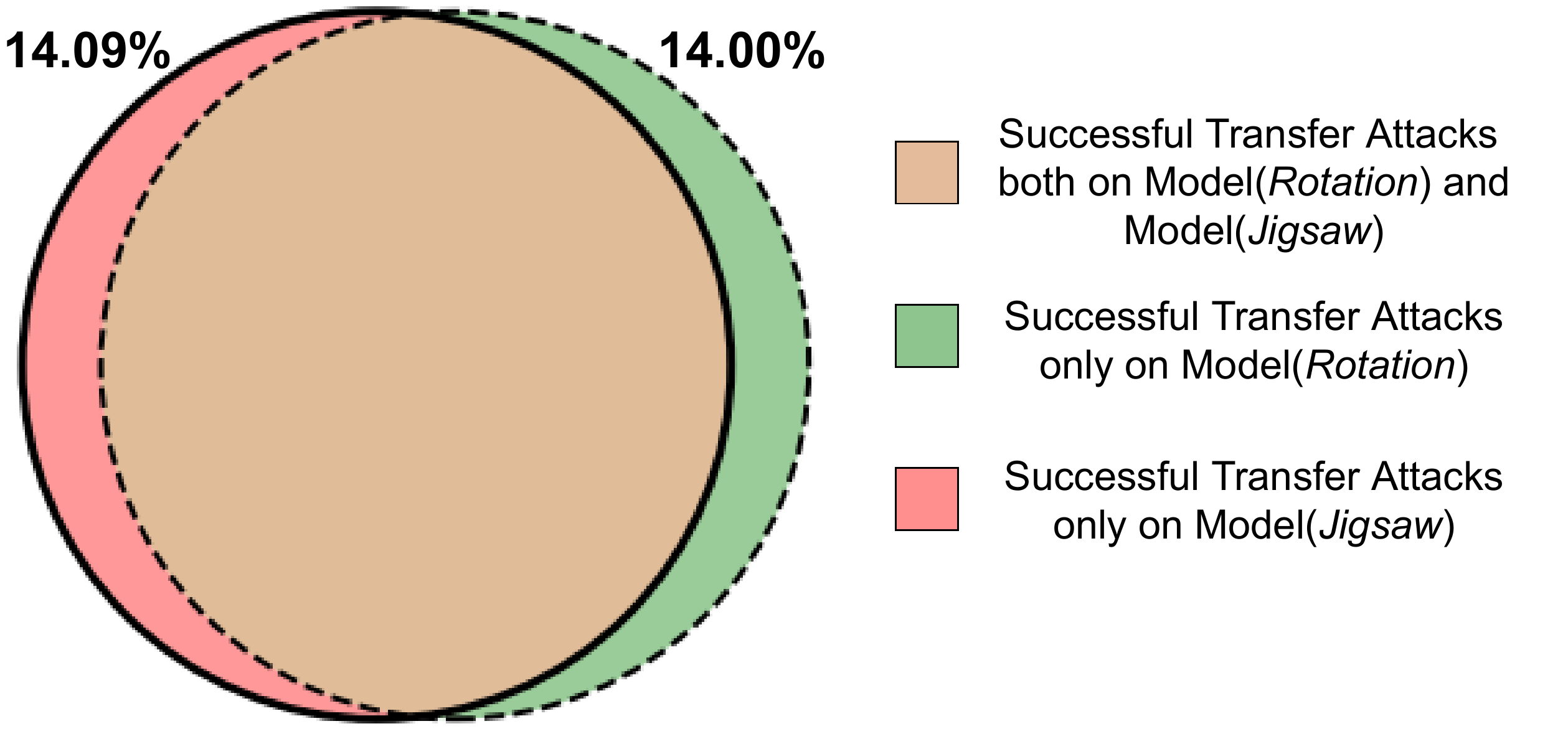}
\end{center}
   \caption{The VENN plot between sets of successful transfer adversarial examples from Model(\textit{Selfie}) to Model(\textit{Rotation}) and Model(\textit{Selfie}) to Model(\textit{Jigsaw}). The overlapping \textit{Brown area} (\textcolor{brown}{\ding{110}}) represents the successful transfer attacks both on Model(\textit{Rotation}) and Model(\textit{Jigsaw}) from Model(\textit{Selfie}). The \textit{Pink area} (\textcolor{Salmon}{\ding{110}}) represents the successful transfer attacks only on Model(\textit{Jigsaw}) from Model\textit{Selfie}. The \textit{Green area} (\textcolor{YellowGreen}{\ding{110}}) represents the successful transfer attacks only on Model(\textit{Rotation}) from Model(\textit{Selfie}).}
\label{fig:diversity}
% \vspace{-3mm}
\end{figure}

At the first glance, the values of ASR of transfer attacks from   $\mathrm{Model}(t)$  to  $\mathrm{Model}(t^\prime)$ ($t^\prime \neq t$) keep similar, e.g., the first column of Table\,\ref{table:diversity} where $t = \textit{Selfie}$  and $t^\prime = \textit{Rotation}$ ($38.92\%$ ASR) or
$t^\prime = \textit{Jigsaw}$ ($38.96\%$ ASR).
 However, Figure\,\ref{fig:diversity} shows that the seemingly similar transferability are built on   more \textit{diverse}   adversarial examples that succeed to attack $\mathrm{Model}(\textit{Rotation})$  and $\mathrm{Model}(\textit{Jigsaw})$, respectively. As we can see, there exist at least $14\%$ transfer examples that are non-overlapped when successfully attacking $\mathrm{Model}(\textit{Rotation})$  and $\mathrm{Model}(\textit{Jigsaw})$. This diverse distribution of transferred adversarial examples against models using different self-supervised pretraining tasks motivates us to further improve the robustness by leveraging an ensemble of diversified pretraining tasks. 
%  our ensemble approach given by\eqref{eq: prob_multiple_self}.
 
In Figure\,\ref{fig:ensemble}, we demonstrate the effectiveness of our proposed adversarial pretraining via diversity-promoted ensemble (AP + DPE)  given in \eqref{eq: prob_multiple_self}.
Here we consider $4$ baseline methods: $3$ single task based adversarial pretraining, and adversarial pretraining via standard ensemble  (AP + SE), corresponding to $\lambda = 0$ in \eqref{eq: prob_multiple_self}. As we can see in Table\,\ref{table:ensemble_pre}, AP + DPE yields at least $1.17\%$ improvement on RA while at most $3.02\%$ degradation on TA, comparing with the best single fine-tuned model. In addition to the ensemble at the pretraining stage, we consider a simple but the most computationally intensive ensemble strategy, an averaged predictions over three final robust models learnt using adversarial pretraining  $\mathcal P_3$ followed by adversarial fine-tuning $\mathcal F_4$ over \textit{Selfie}, \textit{rotation}, and \textit{Jigsaw}.  As we can see in Table\,\ref{table:ensemble_fin}, the best combination, ensemble of three fine-tuned models, yields at least $3.59\%$ on RA while maintains a slight higher TA. More results of other ensemble configurations can be found in the supplement.
 
 % \footnote{ASR is given by the ratio of \textit{successful} adversarial examples over the total number of $10000$ test images.}

% There also exist diversity of different learning tasks in terms of less transferable adversarial examples generated from   task-specific pretraining+fine-tuning classifiers. 
% show that the pretrained embed-dings resulting from different self-supervised tasks havea diverse adversarial vulnerability. In view of that, wepropose to ensemble pretrained embeddings for theircomplementary strengths. On CIFAR-10, our ensemblestrategy further contributes to an improvement of 3.59%on robust accuracy, while maintaining a slightly higherstandard accuracy. T

% The previous results in Table\,\ref{table:p3_matter} and Figure\,\ref{fig:unforeseen} have shown that  the gain of adversarial robustness is consistent and is not very sensitive 

\begin{comment}
\textbf{Competition among Selfie, Rotation and Jigsaw} i) Under scenario ($\mathcal{P}_3$, $\mathcal{F}_1$) and ($\mathcal{P}_3$, $\mathcal{F}_2$), Rotation perform best and Jigsaw perform worse in terms of TA and RA. ii) Under scenario ($\mathcal{P}_3$, $\mathcal{F}_4$), as shown in Figure~\ref{fig:unforeseen}, Selfie consistently perform better than other two tasks, in terms of TA, RA and other 12 unforeseen adversarial attacks excludes ``impluse noise".
\end{comment}

\begin{table}[ht]
\begin{center}
\caption{The vulnerability diversity among fine-tuned models with \textit{Selfie}, \textit{Rotation} and \textit{Jigsaw} self-supervised adversarial pretraining. The results take full adversarial fine-tuning. The highest ASRs are highlighted (\textcolor{red}{$1^{\mathrm{st}}$},\textcolor{blue}{$2^{\mathrm{nd}}$}) under each column of PGD attacks from different fine-tuned models. Ensemble model results to different PGD attacks can be found in our supplement.}
% \vspace{-2.5mm}
\label{table:diversity}
\begin{threeparttable}
\resizebox{0.5\textwidth}{!}{
\begin{tabular}{|@{}c|c|c|c|}
\hline
% \diagbox[trim=l]{Attack}{($\mathcal{P}_3,\mathcal{F}_2$)}{Evaluation} & \begin{tabular}[c]{@{}c@{}}  Model \\ (\textit{Selfie}) \end{tabular}  &  \begin{tabular}[c]{@{}c@{}}  Model \\ (\textit{Rotaion}) \end{tabular}    &\begin{tabular}[c]{@{}c@{}}Model \\(\textit{Jigsaw}) \end{tabular}   \\
% \hline
% \begin{tabular}[c]{@{}c@{}}PGD attacks\\from Model (\textit{Selfie}) \end{tabular} & 62.28\% & 41.84\% & 40.71\%\\
% \hline 
% \begin{tabular}[c]{@{}c@{}}PGD attacks\\from Model (\textit{Rotation}) \end{tabular} & 32.51\% & 54.87\% & 31.86\% \\
% \hline
% \begin{tabular}[c]{@{}c@{}}PGD attacks\\from Model (\textit{Jigsaw}) \end{tabular} & 43.06\% & 43.76\% & 66.32\% \\
% \hline
% \hline
\diagbox[trim=l]{Evaluation}{($\mathcal{P}_3,\mathcal{F}_4$)}{Attack} &\begin{tabular}[c]{@{}c@{}}PGD attacks\\from\\ Model(\textit{Selfie}) \end{tabular} &  \begin{tabular}[c]{@{}c@{}}PGD attacks\\from \\Model(\textit{Rotation}) \end{tabular}   &\begin{tabular}[c]{@{}c@{}}PGD attacks\\from\\ Model(\textit{Jigsaw}) \end{tabular} \\
\hline
\begin{tabular}[c]{@{}c@{}}  Model(\textit{Selfie}) \end{tabular} & \textcolor{red}{48.95\%} & 37.75\% & 36.65\%\\
\hline 
\begin{tabular}[c]{@{}c@{}}  Model(\textit{Rotaion}) \end{tabular} & 38.92\% & \textcolor{red}{49.60\%} & \textcolor{blue}{38.12\%} \\
\hline
\begin{tabular}[c]{@{}c@{}} Model(\textit{Jigsaw}) \end{tabular} & \textcolor{blue}{38.96\%} & \textcolor{blue}{39.56\%} & \textcolor{red}{51.17\%} \\
\hline
\end{tabular}}
\end{threeparttable}
\end{center}
\vspace{-1mm}
\end{table}

\begin{comment}
\textbf{Adversarial Vulnerability Diversity}

% \textbf{Counter Example Visualizations} \TL{TBD}

\subsection{Ensemble self-supervised Tasks}

\textbf{Ensemble Pretraining with Attack Diversity} Our proposed pretraining ensemble strategy in equation (\ref{eq: loss_multiple_self}) achieve the best robust accuracy ($86.04\%$) and a comparable standard accuracy ($86.04\%$).

\end{comment}

\begin{table}[ht]
\begin{center}
\caption{Results comparison between fine-tuned model from single task pretraining and fine-tuned model from tasks ensemble pretraining. AP + SE represents adversarial pretraining via standard ensemble. AP + DPE represents adversarial pretraining via diversity-promoted ensemble. The best results are highlighted (\textcolor{red}{$1^{\mathrm{st}}$},\textcolor{blue}{$2^{\mathrm{nd}}$}) under each column of evaluation metrics.} 
% \vspace{-2.5mm}
\label{table:ensemble_pre}
\begin{threeparttable}
\resizebox{0.47\textwidth}{!}{
\begin{tabular}{c|c|c|c}
\hline
\multirow{1}{*}{Models} & TA (\%)  & RA (\%) & Epochs \\ \hline
\textit{Selfie} Pretraining & \textcolor{red}{86.02} & \textcolor{blue}{51.05} & 50 \\
\textit{Rotation} Pretraining & \textcolor{blue}{85.66} & 50.40 & \textcolor{red}{46} \\
\textit{Jigsaw} Pretraining & 83.74 & 48.83 & 48 \\
\hline
\hline
\rowcolor[gray]{0.9} 
AP + SE & 84.44 & 49.53 & \textcolor{blue}{47} \\
\rowcolor[gray]{0.9} 
AP + DPE & 83.00 & \textcolor{red}{52.22} & 56 \\
\hline
\hline
\end{tabular}}
\end{threeparttable}
\end{center}
\vspace{-3mm}
\end{table}

% \begin{figure}[ht]
% \begin{center}
%   \includegraphics[width=1\linewidth]{Pics/ensemble_pre.png}
% \end{center}
%   \caption{Performance of Different Pretraining Ensemble Strategies}
% \label{fig:ensemble_pre}
% \vspace{-3mm}
% \end{figure}

% \TL{Mention the details are in the supplement.} TA = 81.8 and RA = 52.86, $\sim$ 110 epochs $\gamma$ = 0.05

\begin{comment}
\textbf{Ensumble Fine-tuning with Adaptive Diversity Promoting (ADP) Regularizer}

\textbf{Ensmeble Fine-tuned Models} i) Notice that the best performance achieve by single model under scenario ($\mathcal{P}_3$, $\mathcal{F}_4$) is Selfie's TA: 86.02 and RA: 51.05. In Table~\ref{table:ensemble_fin}, our ensemble fine-tuned model surpass previous best RA by 3.59 points. ii) From the performance of three combinations in Table~\ref{table:ensemble_fin}, it also revel \textit{Selfie} is the most beneficial task under scenario ($\mathcal{P}_3$, $\mathcal{F}_4$) and \textit{Jigsaw} is the worst.

\end{comment}

\begin{table}[ht]
\begin{center}
\caption{Ensemble results of fine-tuned models with different adversarial pretrainings. The best results are highlighted (\textcolor{red}{$1^{\mathrm{st}}$},\textcolor{blue}{$2^{\mathrm{nd}}$}) under each column of evaluation metrics.} 
% \vspace{-2.5mm}
\label{table:ensemble_fin}
\begin{threeparttable}
\resizebox{0.47\textwidth}{!}{
\begin{tabular}{c|c|c}
\hline
\multirow{1}{*}{Fine-tuned Models ($\mathcal{P}_3,\mathcal{F}_4$)} & TA (\%)  & RA (\%) \\ \hline
\textit{Jigsaw} + \textit{Rotation} & 85.36 & 53.08 \\
\textit{Jigsaw} + \textit{Selfie} & 85.64 & 53.32 \\
\textit{Rotation} + \textit{Selfie} & \textcolor{red}{86.51} & \textcolor{blue}{53.83} \\
\hline
\hline
\rowcolor[gray]{0.9} 
\textit{Jigsaw} + \textit{Rotation} + \textit{Selfie} & \textcolor{blue}{86.04} & \textcolor{red}{54.64} \\
\hline
\hline
\end{tabular}}
\end{threeparttable}
\end{center}
\vspace{-3mm}
\end{table}

\subsection{Ablation Study and Analysis} \label{sec: abla}
For comparison fairness, we fine-tune all models in the same CIFAR-10 dataset. In each ablation, we show results under scenarios ($\mathcal{P}_3,\mathcal{F}_2$) and ($\mathcal{P}_3,\mathcal{F}_4$), where $\mathcal{P}_3$ represents adversarial pretraining, $\mathcal{F}_2$ represents partial adversarial fine-tuning and $\mathcal{F}_4$ represents full adversarial fine-tuning. More ablation results can be found in the supplement.

% \begin{table}[ht]
% \begin{center}
% \caption{Ablation results of the size of pretraining datasets. All pretraining datasets have $32\times32$ resolution and $10$ classes.} 
% % \vspace{-2.5mm}
% \label{table:aba1}
% \begin{threeparttable}
% \resizebox{0.48\textwidth}{!}{
% \begin{tabular}{c|c|c|c|c|c|c|c|c|c}
% \hline
% \multirow{2}{*}{Scenario} & \multicolumn{3}{c|}{CIFAR-30K} & \multicolumn{3}{c|}{CIFAR-50K} & \multicolumn{3}{c}{CIFAR-150K}\\ \cline{2-10} 
%  & TA (\%) & RA (\%) & Epochs & TA (\%)  & RA (\%) & Epochs & TA (\%) & RA (\%) & Epochs\\ \hline
% ($\mathcal{P}_3,\mathcal{F}_2$) & 65.65 & 30.00 & 70 & 66.87 & 30.42 & 87 & 67.73 & 30.24 & 95\\
% ($\mathcal{P}_3,\mathcal{F}_4$) & 85.29 & 49.64 & 42 & 85.26 & 49.66 & 61 & 85.18 & 50.61 & 55\\
% \hline
% \hline
% \end{tabular}}
% \end{threeparttable}
% \end{center}
% \vspace{-1mm}
% \end{table}

\begin{table}[ht]
\begin{center}
\caption{Ablation results of the size of pretraining datasets. All pretraining datasets have $32\times32$ resolution and $10$ classes.} 
% \vspace{-2.5mm}
\label{table:aba1}
\begin{threeparttable}
\resizebox{0.4\textwidth}{!}{
\begin{tabular}{c|c|c|c}
\hline
\multirow{2}{*}{Scenario} & \multicolumn{3}{c}{CIFAR-30K} \\ \cline{2-4} 
 & TA (\%) & RA (\%) & Epochs\\ \hline
($\mathcal{P}_3,\mathcal{F}_2$) & 65.65 & 30.00 & 70 \\
($\mathcal{P}_3,\mathcal{F}_4$) & 85.29 & 49.64 & 42 \\
\hline
\hline
\multirow{2}{*}{Scenario} & \multicolumn{3}{c}{CIFAR-50K} \\ \cline{2-4} 
& TA (\%)  & RA (\%) & Epochs\\ \hline
($\mathcal{P}_3,\mathcal{F}_2$) &  66.87 & 30.42 & 87\\
($\mathcal{P}_3,\mathcal{F}_4$) & 85.26 & 49.66 & 61\\
\hline
\multirow{2}{*}{Scenario} & \multicolumn{3}{c}{CIFAR-150K}\\ \cline{2-4} 
& TA (\%)  & RA (\%) & Epochs \\ \hline
($\mathcal{P}_3,\mathcal{F}_2$)  & 67.73 & 30.24 & 95\\
($\mathcal{P}_3,\mathcal{F}_4$) & 85.18 & 50.61 & 55\\
\hline
\end{tabular}}
\end{threeparttable}
\end{center}
\vspace{-3mm}
\end{table}

% \begin{table}[ht]
% \begin{center}
% \caption{Ablation results of the number of classes in the pretraining datasets. Both pretraining datasets have $32\times32$ resolution.} 
% % \vspace{-2.5mm}
% \label{table:aba2}
% \begin{threeparttable}
% \resizebox{0.46\textwidth}{!}{
% \begin{tabular}{c|c|c|c|c|c|c}
% \hline
% \multirow{2}{*}{Scenario} & \multicolumn{3}{c|}{CIFAR-10 (10 classes)} & \multicolumn{3}{c}{ImageNet-32 (1000 classes)} \\ \cline{2-7} 
%  & TA (\%) & RA (\%) & Epochs  & TA (\%) & RA (\%) & Epochs \\ \hline
% % ($\mathcal{P}_3,\mathcal{F}_1$) & 78.93 & 6.30 & 82 & 80.4  & 3.47 & 95 \\
% ($\mathcal{P}_3,\mathcal{F}_2$) & 74.30 & 37.65 & 64 & 72.96 & 36.40 & 76 \\
% ($\mathcal{P}_3,\mathcal{F}_4$) & 86.02 & 51.05 & 50 & 85.56 & 51.10 & 44 \\
% \hline
% \hline
% \end{tabular}}
% \end{threeparttable}
% \end{center}
% \vspace{-3mm}
% \end{table}

\begin{table}[ht]
\begin{center}
\caption{Ablation results of defense approaches. Instead of adversarial training, we perform random smoothing \cite{cohen2019certified} for pretraining.} 
% \vspace{-2.5mm}
\label{table:random}
\begin{threeparttable}
\resizebox{0.48\textwidth}{!}{
\begin{tabular}{c|c|c|c|c|c|c|c|c|c}
\hline
\multirow{2}{*}{\begin{tabular}[c]{@{}c@{}}Random \\ Smoothing \end{tabular}} & \multicolumn{3}{c|}{\textit{Selfie} Pretrainng} & \multicolumn{3}{c|}{\textit{Rotation} Pretraining} & \multicolumn{3}{c}{\textit{Jigsaw} Pretraining}\\ \cline{2-10} 
 & TA (\%) & RA (\%) & Epochs & TA (\%) & RA (\%) & Epochs & TA (\%) & RA (\%) & Epochs\\ \hline
$\mathcal{F}_2$ & 71.9 & 30.57 & 61 & 74.7 & 34.23 & 78 & 74.66 & 33.84 & 68\\
$\mathcal{F}_4$ & 85.14 & 50.23 & 48 & 85.62 & 51.25 & 46 & 85.18 & 50.94 & 46\\
\hline
\hline
\end{tabular}}
\end{threeparttable}
\end{center}
\vspace{-5mm}
\end{table}

\paragraph{Ablation of the pretraining data size} As shown in Table\,\ref{table:aba1}, as the pretraining dataset grows larger, the standard and robust accuracies both demonstrate steady growth. Under the ($\mathcal{P}_3,\mathcal{F}_4$) scenario, when the pretraining data size increases from 30K to 150K, we observe a 0.97\% gain on robust accuracy with nearly the same standard accuracy. That aligns with the existing theory \cite{schmidt2018adversarially}. Since self-supervised pretraining requires no label, we could in future grow the unlabeled data size almost for free to continuously boost the pretraining performance.

\paragraph{Ablation of defense approaches in pretraining} In Table\,\ref{table:random}, we use random smoothing \cite{cohen2019certified} in place of AT to robustify pretraining, while other protocols remain all unchanged. We obtain consistent results to using adversarial pretraining: robust pretraining speed up adversarial fine-tuning and helps final model robustness, while the full adversarial fine-tuning contributes the most to the robustness boost. 

\section{Conclusions}
\vspace{-2mm}
In this paper, we combine adversarial training with self-supervision to gain robust pretrained models, that can be readily applied towards downstream tasks through fine-tuning. We find that adversarial pretraining can not only boost final model robustness but also speed up the subsequent adversarial fine-tuning. We also find adversarial fine-tuning to contribute the most to the final robustness improvement. Further motivated by our observed diversity among different self-supervised tasks in pretraining, we propose an ensemble pretraining strategy that boosts robustness further. Our results observe consistent gains over state-of-the-art AT in terms of both standard and robust accuracy, leading to new benchmark numbers on CIFAR-10. In the future, we are interested to explore several promising directions revealed by our experiments and ablation studies, including incorporating more self-supervised tasks, extending the pretraining dataset size, and scaling up to high-resolution data.

{\small
\bibliographystyle{ieee_fullname}
\bibliography{egbib}
}

%\clearpage
%\newpage
% \appendix

\end{document}